\documentclass[pdflatex, sn-mathphys]{sn-jnl}

\usepackage{amsmath}
\usepackage{amssymb}
\usepackage{multirow}
\usepackage{dsfont}
\usepackage{hyperref}
\usepackage{epsfig}
\usepackage{graphicx}
\usepackage{xcolor}
\usepackage{diagbox}
\usepackage{xspace}
\usepackage{tabularx,booktabs}
\usepackage{url}
\usepackage{caption}
\usepackage{subcaption}
\usepackage{hhline}

\usepackage[nice]{nicefrac}

\newcommand{\xx}{\mathbf{x}\xspace}
\newcommand{\xt}{\mathbf{\widetilde{x}}\xspace}

\jyear{2021}%

\theoremstyle{thmstyleone}%
\theoremstyle{thmstyletwo}%

\theoremstyle{thmstylethree}%

\begin{document}
\newcommand{\figVQGAN}{
\begin{figure*}[t]
\centering
\includegraphics[width=1.0\linewidth]{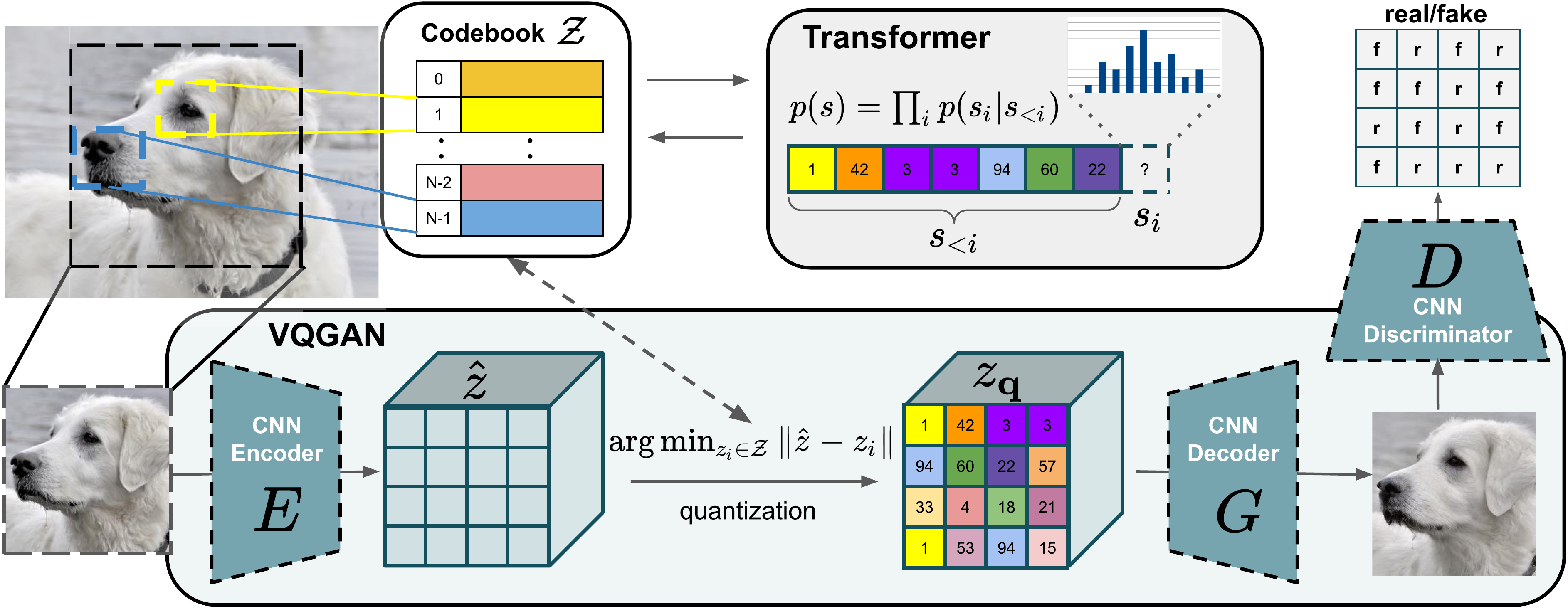} \\
\caption{Overview of VQ-GAN~\cite{Esser2021TamingTF}.}
\label{fig:VQGAN}
\end{figure*}
}

\newcommand{\figSemanticStyleGAN}{
\begin{figure*}[t]
\centering
\includegraphics[width=1.0\linewidth]{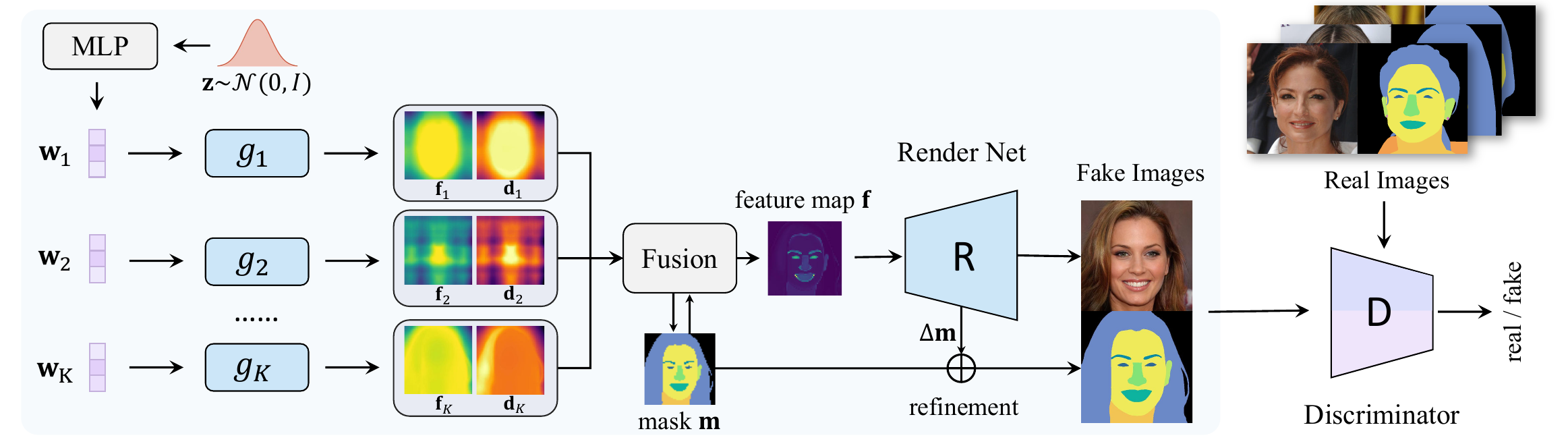} \\
\caption{Overview of SemanticStyleGAN~\cite{Shi.2022}.}
\label{fig:SemanticStyleGAN}
\end{figure*}
}

\newcommand{\figGANDissection}{
\begin{figure}
    \centering
    \includegraphics[width=0.9\textwidth]{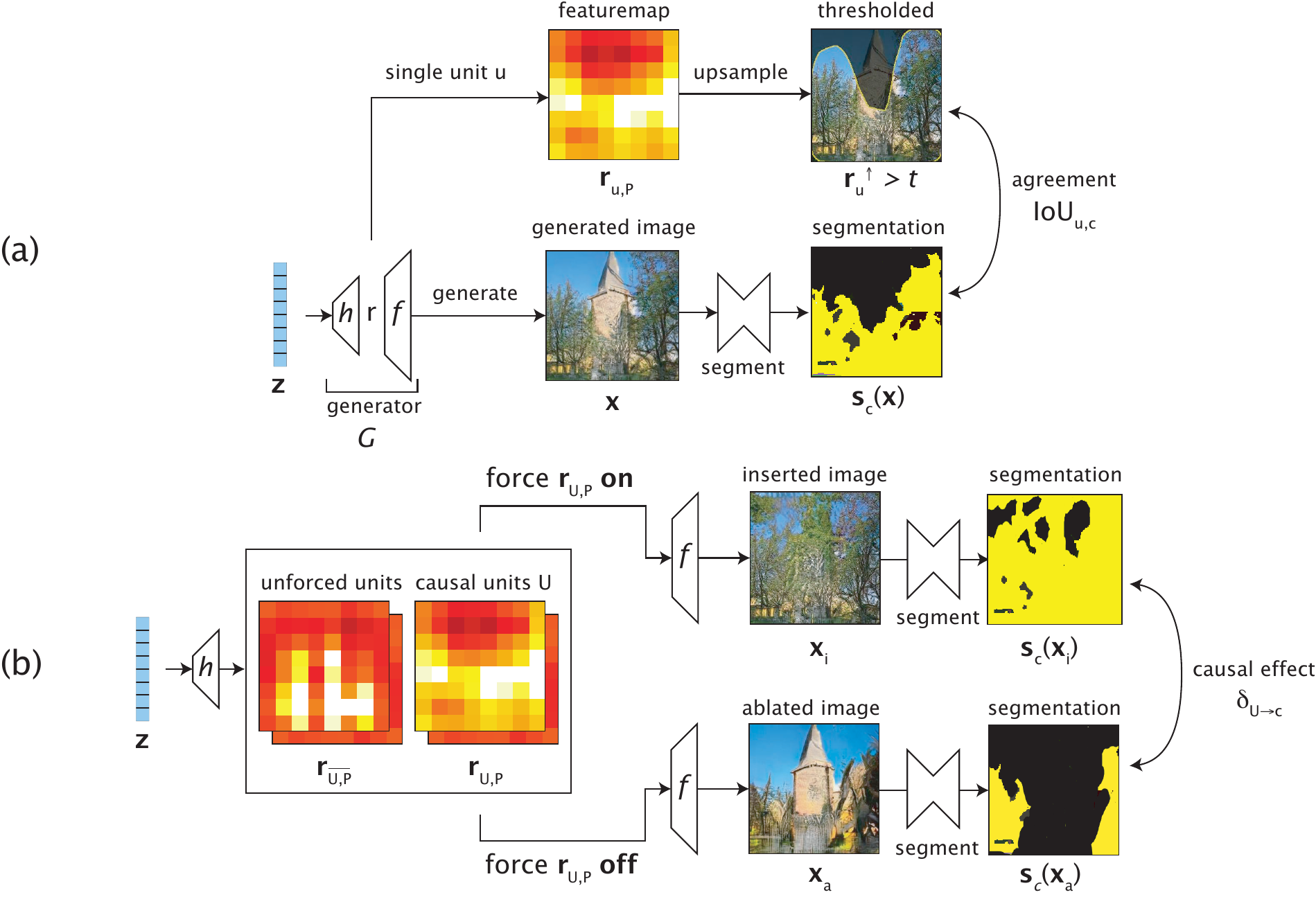}
    \caption{Overview of the method presented in GAN Dissection~\cite{Bau.2018}.}
    \label{fig:GAN-Dissection}
\end{figure}
}

\newcommand{\figStyleGANs}{
\begin{figure}\centering
    \begin{subfigure}[b]{0.6\textwidth}
        \includegraphics[width=0.90\textwidth]{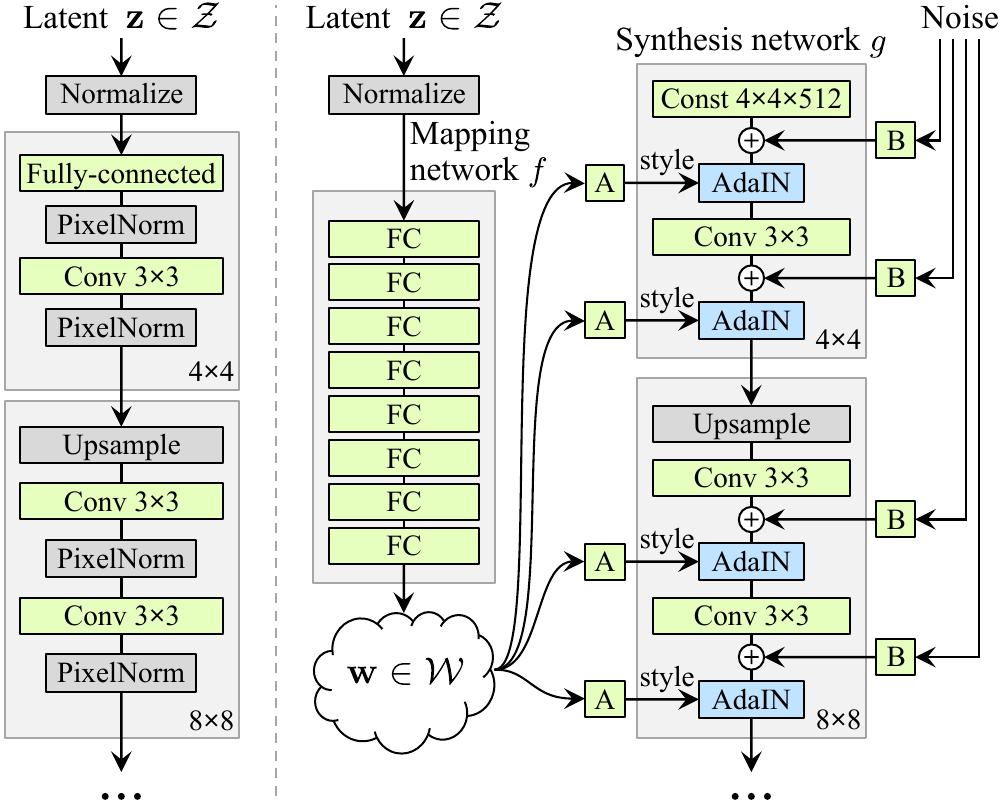} 
        \caption{}
        \label{fig:stylegan1_arch}
    \end{subfigure}
    \hfill
    \begin{subfigure}[b]{0.35\textwidth}
        \includegraphics[width=0.8\textwidth]{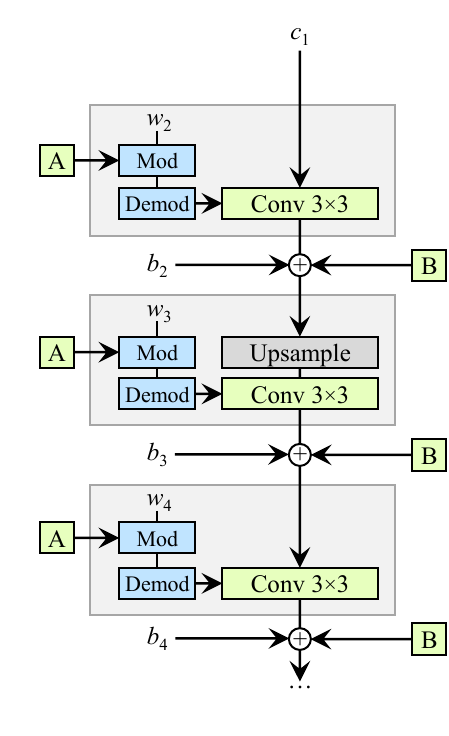}
        \caption{}
        \label{fig:stylegan2_arch}
    \end{subfigure} 
    \caption{StyleGAN architecture. (a) the structure of the StyleGAN~\cite{Karras_2019_CVPR} generator. (b) a focus on the improved synthesis network of StyleGAN2~\cite{Karras_2020_CVPR}}
    \label{fig:stylegan1_2_arch}
\end{figure}
}

\newcommand{\figsBigGAN}{
\begin{figure}\centering
    \begin{subfigure}[b]{0.3\textwidth}
        \includegraphics[width=\textwidth]{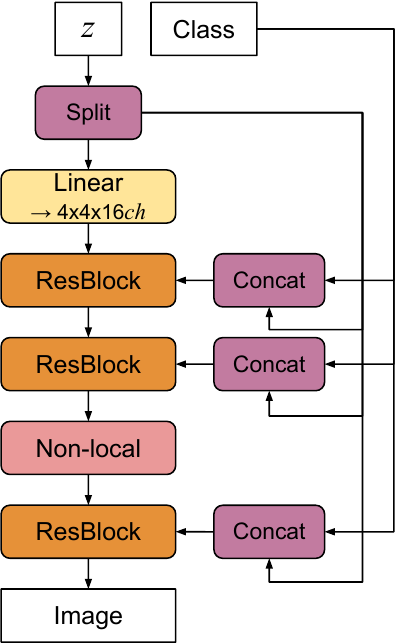}
        \caption{}
        \label{fig:biggan1}
    \end{subfigure}
    \hfill
        \begin{subfigure}[b]{0.3\textwidth}
        \includegraphics[width=\textwidth]{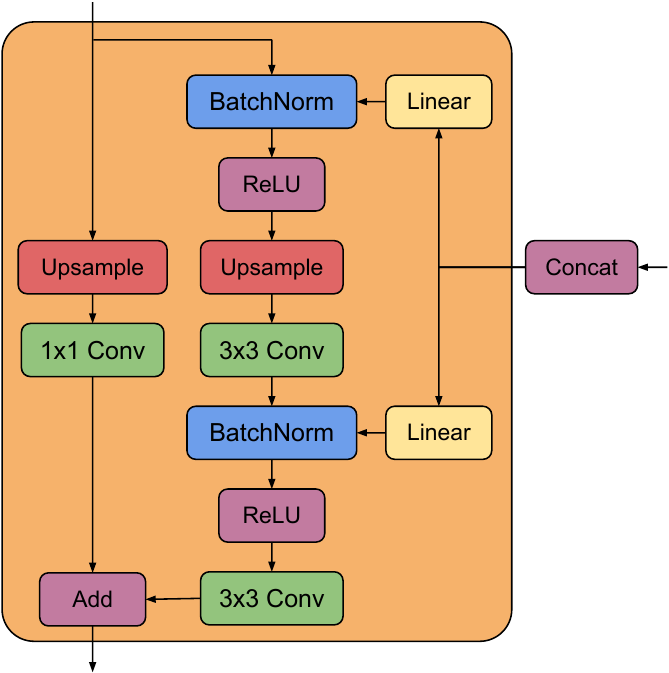}
        \caption{}
        \label{fig:biggan2}
    \end{subfigure}
    \hfill
        \begin{subfigure}[b]{0.3\textwidth}
        \includegraphics[width=\textwidth]{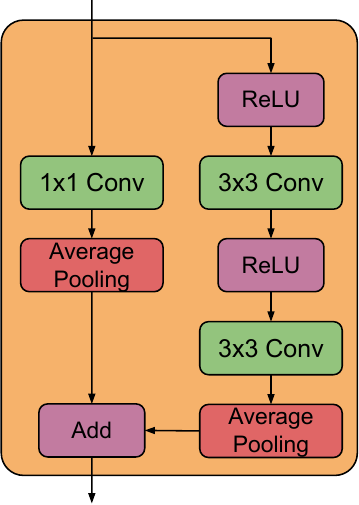}
        \caption{}
        \label{fig:biggan3}
    \end{subfigure}
     \hfill
    \caption{BigGAN architecture~\cite{Brock.25.02.2019}. (a) Generator layout. (b) Generator residual block. (c) Discriminator residual block.}
    \label{fig:BigGAN_arch}
\end{figure}
}

\newcommand{\figSpadeSean}{%
\begin{figure}
    \centering
\begin{tabular}{cc}
\begin{subfigure}[b]{0.3\textwidth}
        \includegraphics[width=\textwidth]{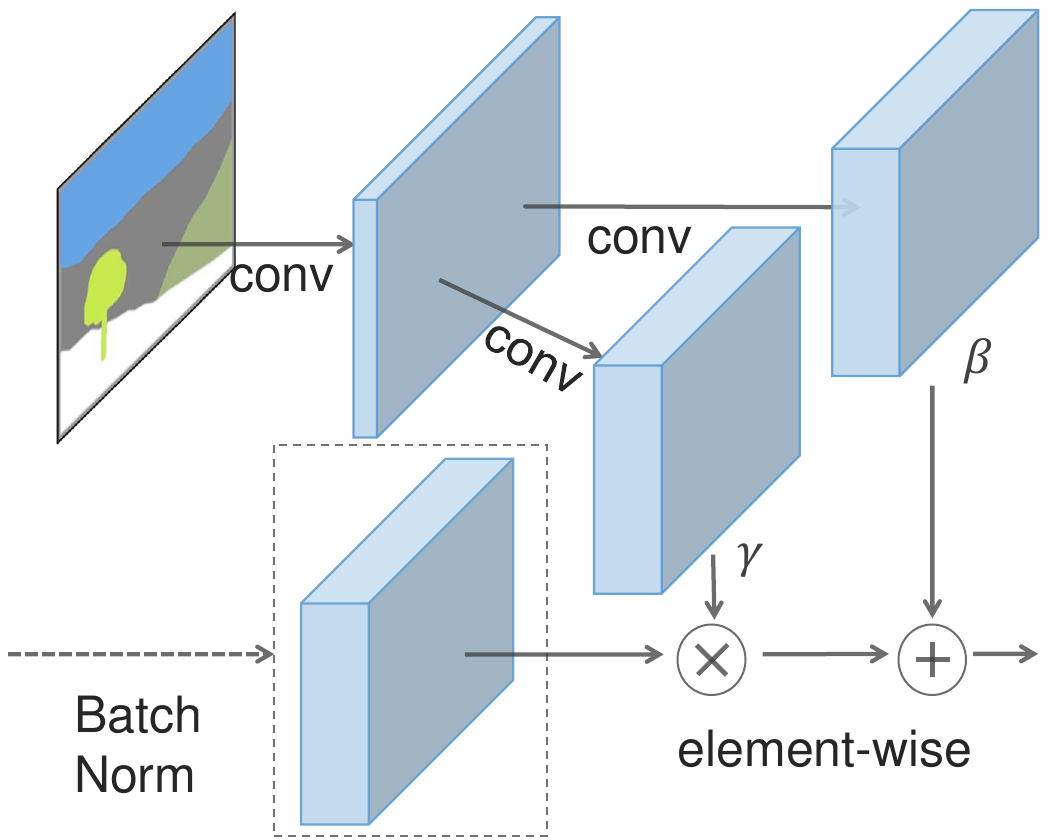}
        \caption{}
        \label{fig:spade_spatial_ada_morm}
    \end{subfigure}
 & 
 \begin{subfigure}[b]{0.3\textwidth}
        \includegraphics[width=\textwidth]{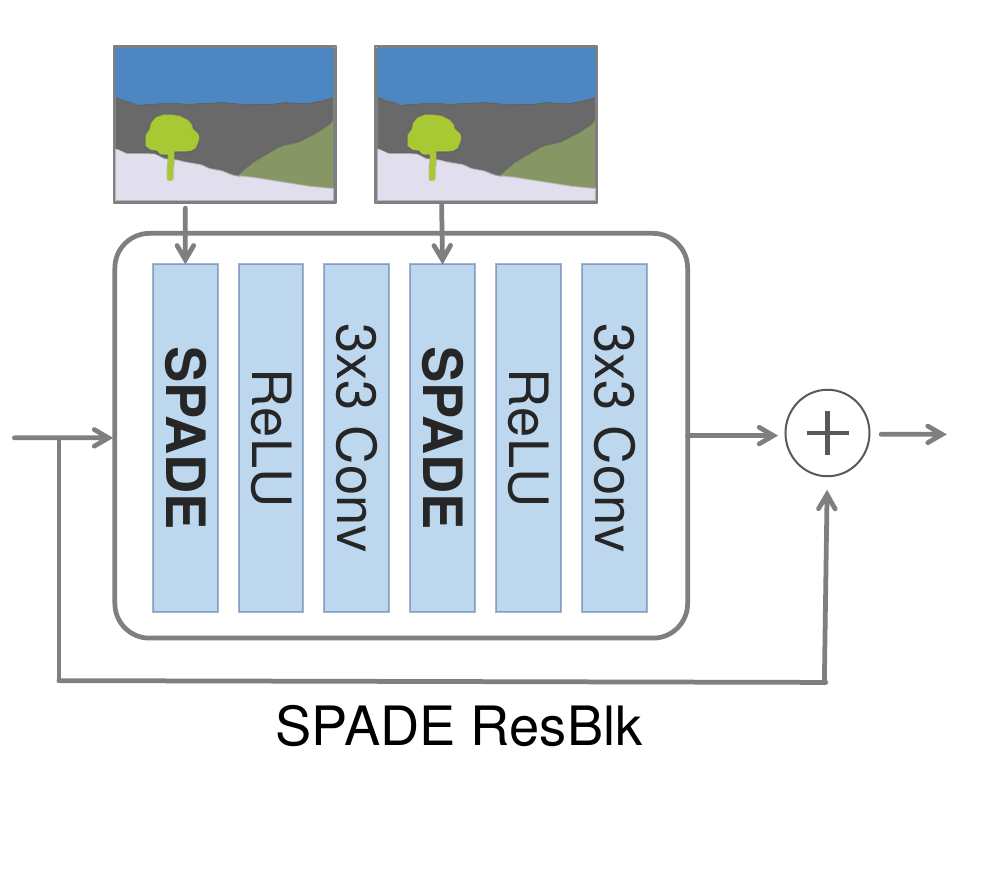}
        \caption{}
        \label{fig:spade_resblk}
    \end{subfigure}
  \\
 \begin{subfigure}[b]{0.5\textwidth}
        \includegraphics[width=\textwidth]{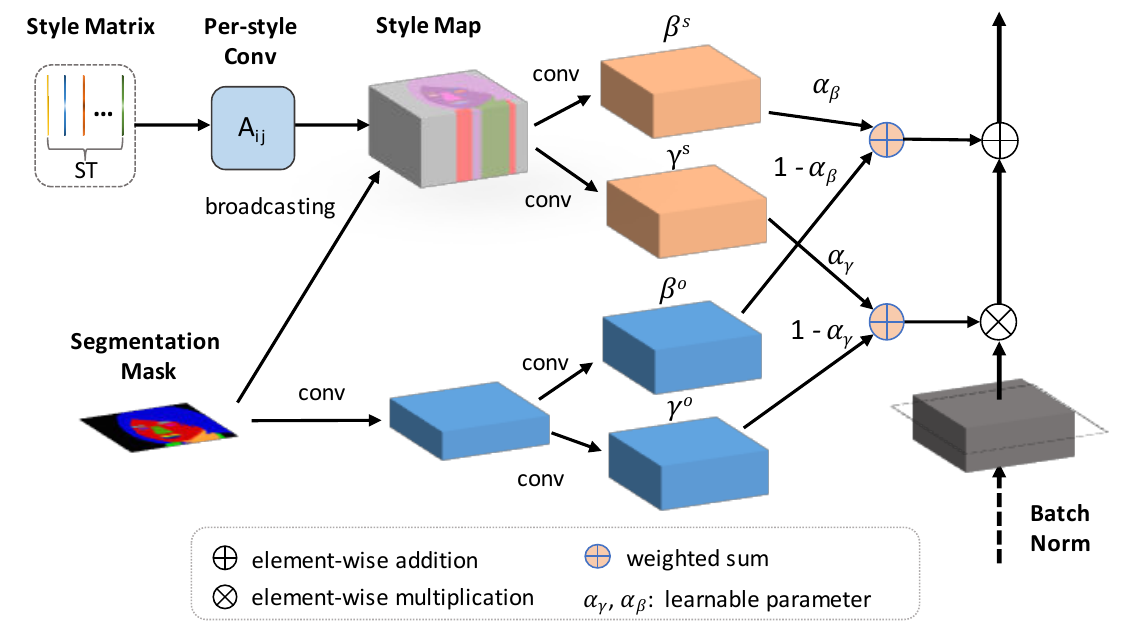}
        \caption{}
        \label{fig:sean_norm}
    \end{subfigure}
    & 
     \begin{subfigure}[b]{0.5\textwidth}
        \includegraphics[trim=0 0 0 0.5\textwidth ,width=\textwidth]{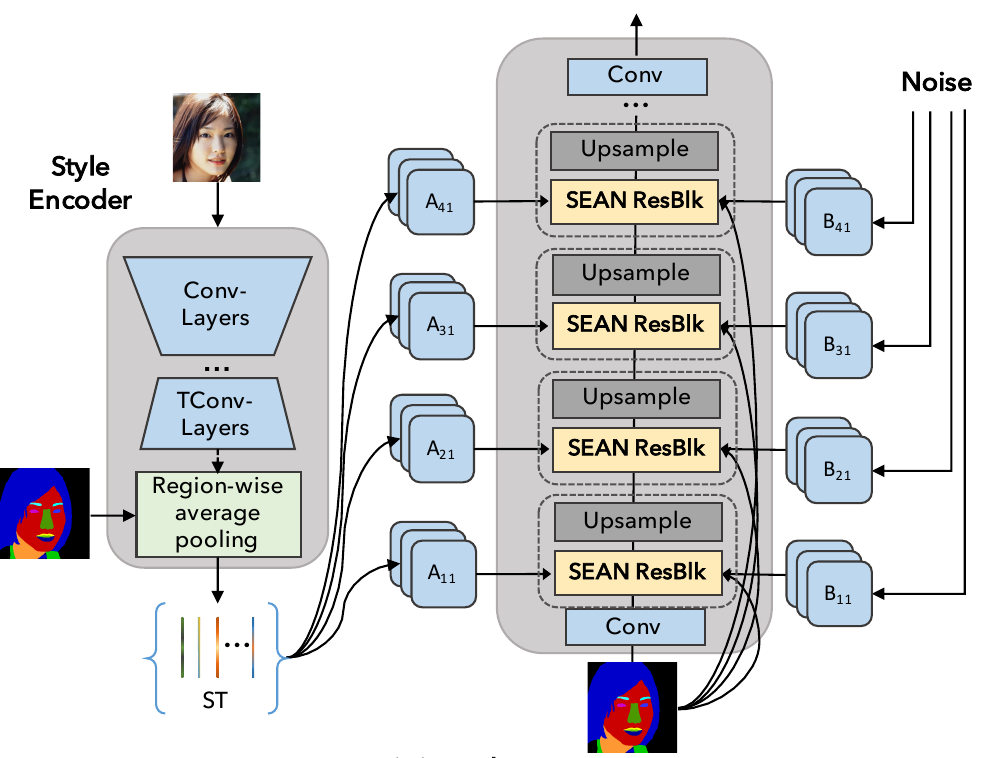}
        \caption{}
        \label{fig:sean_pipeline}
    \end{subfigure}
     \\
\end{tabular}
\caption{Overview of SPADE~\cite{park2019semantic} and SEAN~\cite{zhu2020sean}: (a) Spatially-adaptive normalization of SPADE,  (b) SPADE ResBlk, (c) SEAN ResBlk and (d) SEAN pipeline.}
\end{figure}
}

\newcommand{\figFewShotGenerativeModelAdaptation}{
\begin{figure}[t]
    \centering
    \includegraphics[width=0.9\textwidth]{figs/few_shot_generative_model_adaptation_pipeline.pdf}
    \caption{The framework of the method presented in \cite{Xiao.2022}.}
    \label{fig:GAN-Dissection}
\end{figure}
}

\newcommand{\figSeeingWhatsGAN}{
\begin{figure}[t]
    \centering
    \includegraphics[width=0.6\textwidth]{figs/Seeing_what_GAN_framework-crop.pdf}
    \caption{Overview of the layer inversion method presented in \cite{Bau.2019}.}
    \label{fig:SeeingWhatsGAN}
\end{figure}
}

\newcommand{\figInDomainGAN}{
\begin{figure}[t]
    \centering
    \includegraphics[width=0.6\textwidth]{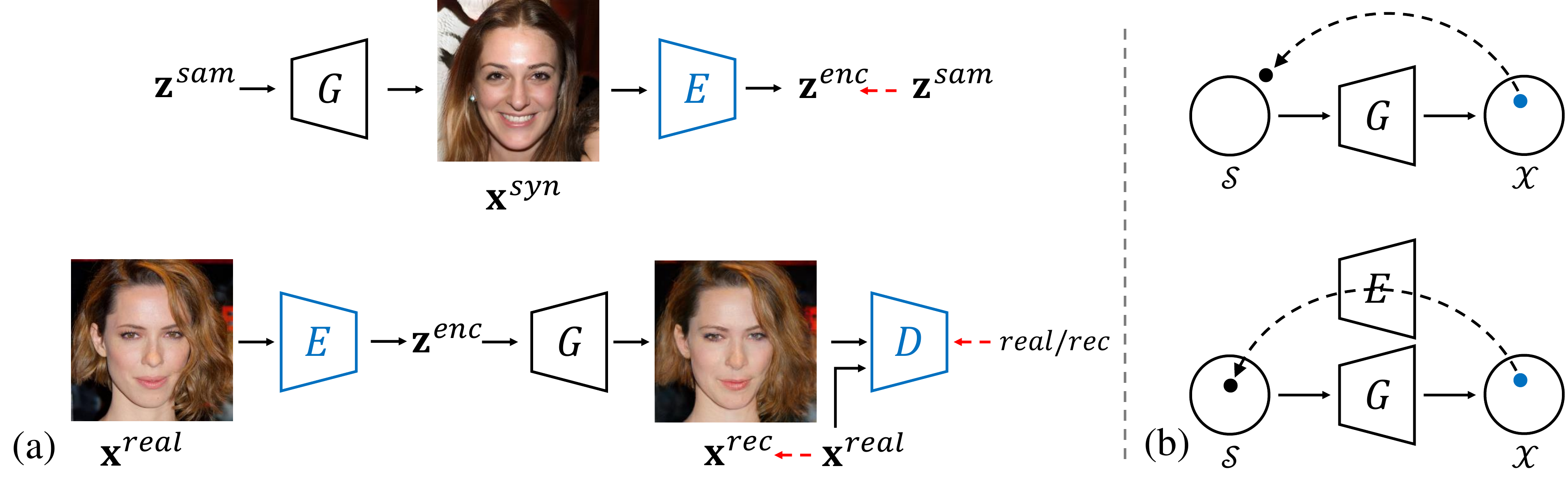}
    \caption{The overview of In-Domain GAN Inversion~\cite{Zhu.2020}. (a) The comparison between the training of a conventional encoder and a domain-guided encoder for GAN inversion. (b) The comparison between the conventional optimization and a domain-regularized
optimization.}
    \label{fig:inDomainGAN}
\end{figure}
}

\newcommand{\figpSp}{
\begin{figure}[t]
    \centering
    \includegraphics[width=0.9\textwidth]{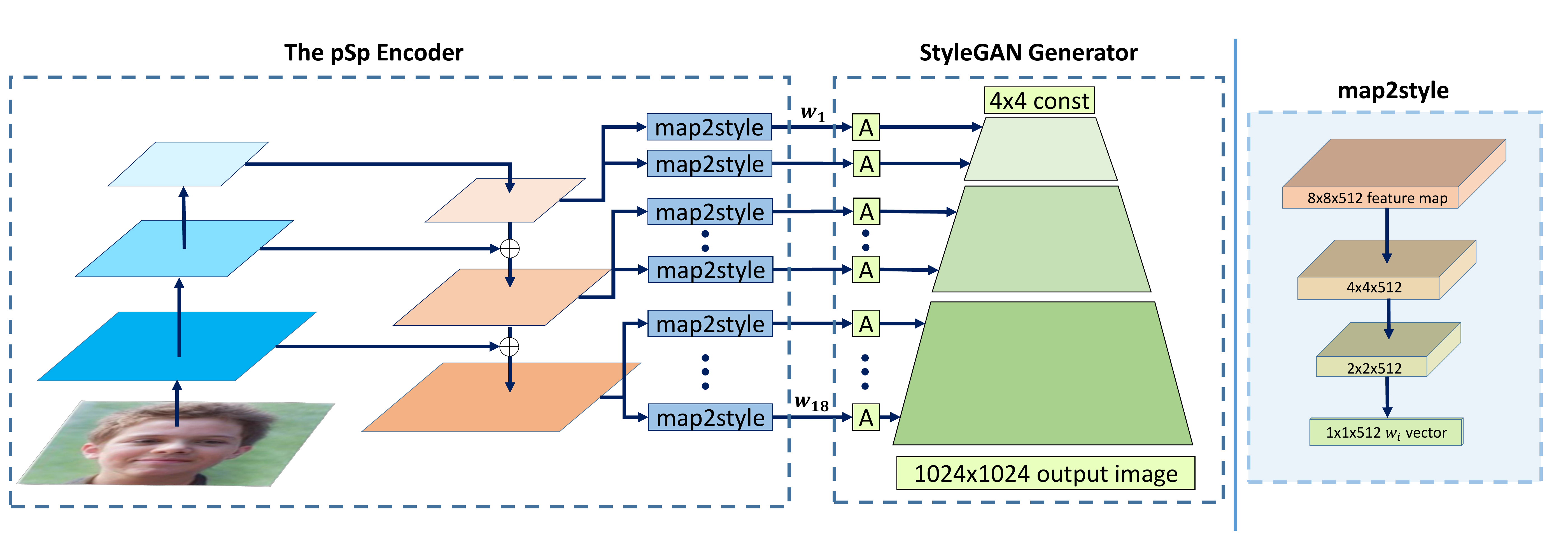}
    \caption{The pSp architecture~\cite{richardson2021encoding}}
    \label{fig:pSp}
\end{figure}
}

\newcommand{\figptpHD}{
\begin{figure*}[t]
\centering
\includegraphics[width=1.0\linewidth]{figs/pix2pixHD_arch.pdf} \\
\caption{The generator architecture of pix2pixHD. }
\label{fig:ptpHD}
\end{figure*}
}

\newcommand{\figcycleGAN}{
\begin{figure*}[t]
\centering
\includegraphics[width=1.0\linewidth]{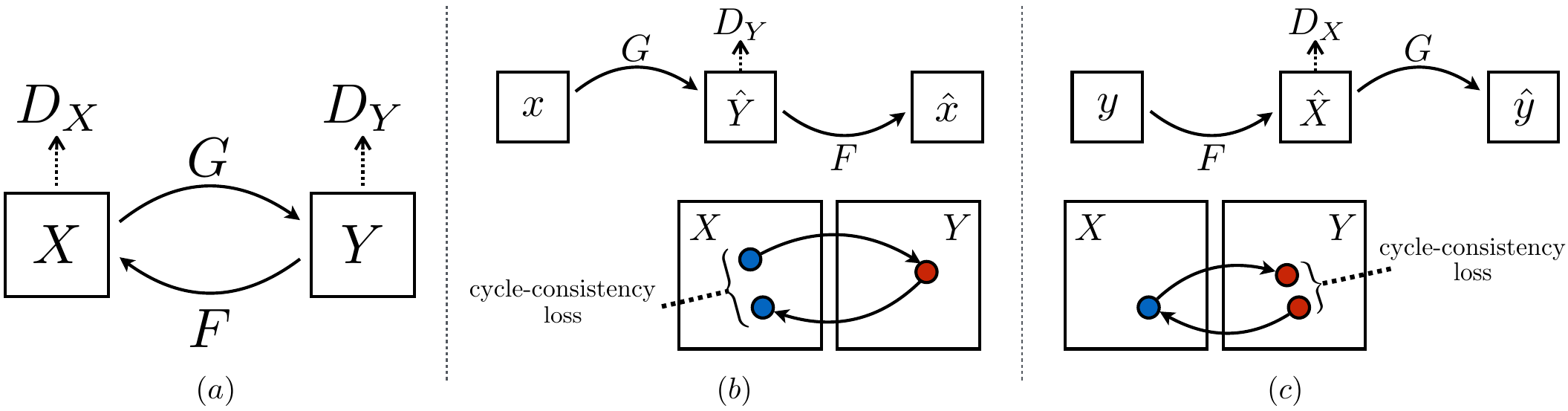} \\
\caption{The overview of CycleGAN~\cite{Zhu_2017_ICCV}. (a) Two domain $X$ and $Y$ are connected via two mapping functions $G$ and $F$. (b) Forward cycle-consistency. (c) Backward cycle-consistency.}
\label{fig:cycleGAN}
\end{figure*}
}

\newcommand{\figMUNIT}{
\begin{figure*}[t]
\centering
\includegraphics[width=1.0\linewidth]{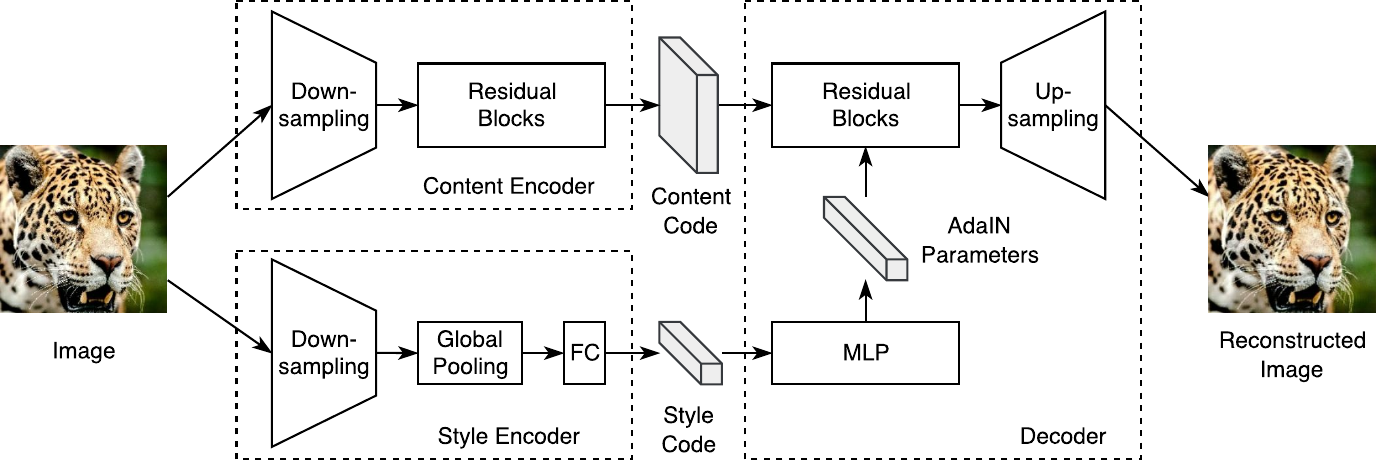} \\
\caption{The auto-encoder architecture of MUNIT~\cite{Huang2018MultimodalUI}.}
\label{fig:MUNIT}
\end{figure*}
}

\newcommand{\figStarGANtwo}{
\begin{figure*}[t]
\centering
\includegraphics[width=1.0\linewidth]{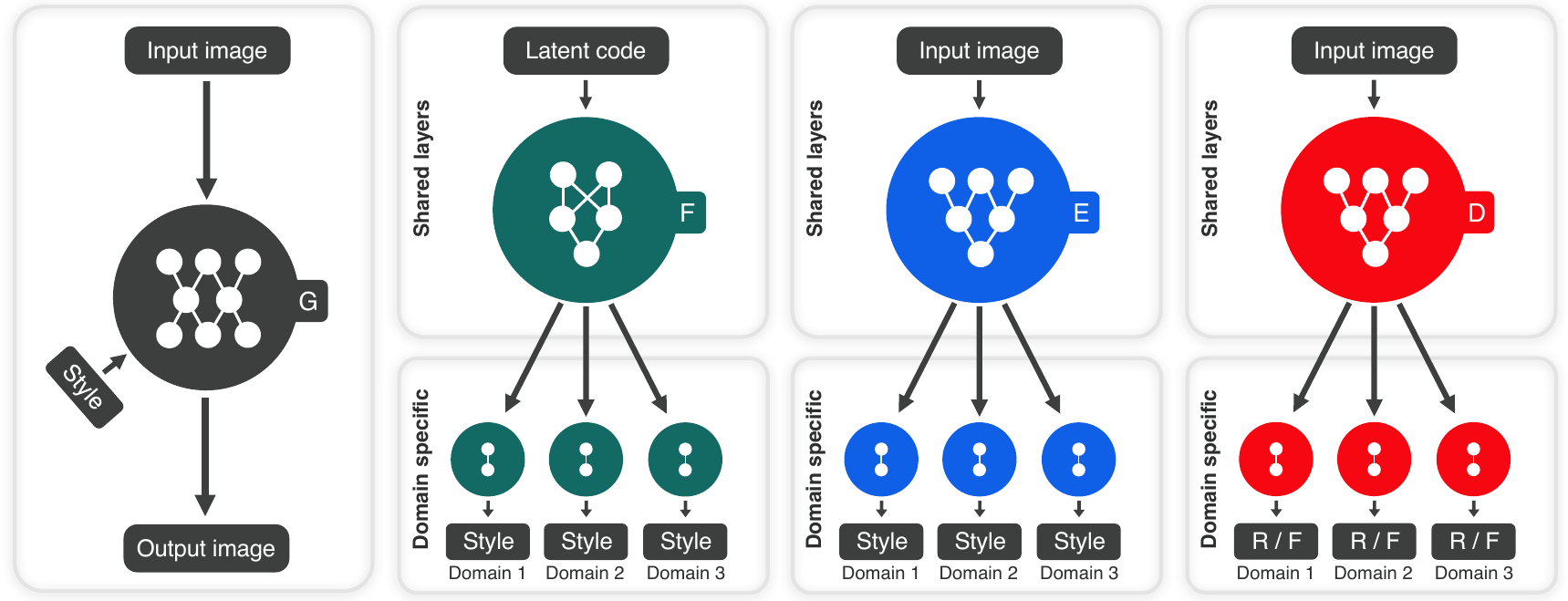} \\
\caption{The Overview of StarGAN v2~\citep{Choi_2018_CVPR}.}
\label{fig:starganv2}
\end{figure*}
}

\newcommand{\figImgAug}{
\begin{figure}[tb]
\centering
\begin{subfigure}{0.162\linewidth}
\frame{\includegraphics[width=\linewidth]{figs/aug/dog.png}}
\caption*{Original}\end{subfigure}  
\begin{subfigure}{0.162\linewidth}
\frame{\includegraphics[width=\linewidth]{figs/aug/zoomin.png}}
\caption*{ZoomIn}\end{subfigure}  
\begin{subfigure}{0.162\linewidth}
\frame{\includegraphics[width=\linewidth]{figs/aug/translation.png}}
\caption*{Translation}\end{subfigure}  
\begin{subfigure}{0.162\linewidth}
\frame{\includegraphics[width=\linewidth]{figs/aug/instancenoise.png}}
\caption*{InstanceNoise}\end{subfigure}  
\begin{subfigure}{0.162\linewidth}
\frame{\includegraphics[width=\linewidth]{figs/aug/cutmix.png}}
\caption*{CutMix\cite{CUTMIX}}\end{subfigure}  
\caption{Examples of augmentation techniques applied to the original image in \cite{Zhao2020ImageAF}.
}
\label{fig:ImgAug}
\end{figure}
}

\newcommand{\figDiffAug}{
\begin{figure*}[t]
\centering
\includegraphics[width=1.0\linewidth]{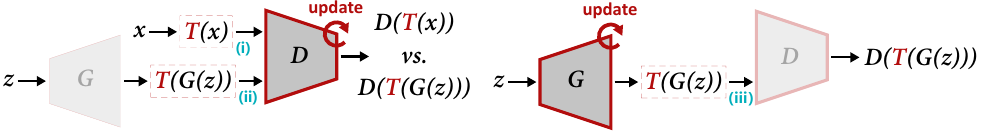} \\
\caption{Overview of DiffAugment~\cite{zhao2020differentiable} for updating $D$ (left) and $G$ (right). }
\label{fig:diffaug}
\end{figure*}
}

\newcommand{\figADA}{
\begin{figure*}[t]
\footnotesize%
\centering%
\newcommand{\h}{38mm}%
\newcommand{\hh}{\h/\real{14}*\real{9.05}}%
\newcommand{\hhh}{10.7mm}%
\newcommand{\vv}{27.8mm}%
\parbox[b][\vv]{\h}{%
\includegraphics[width=1.0\linewidth]{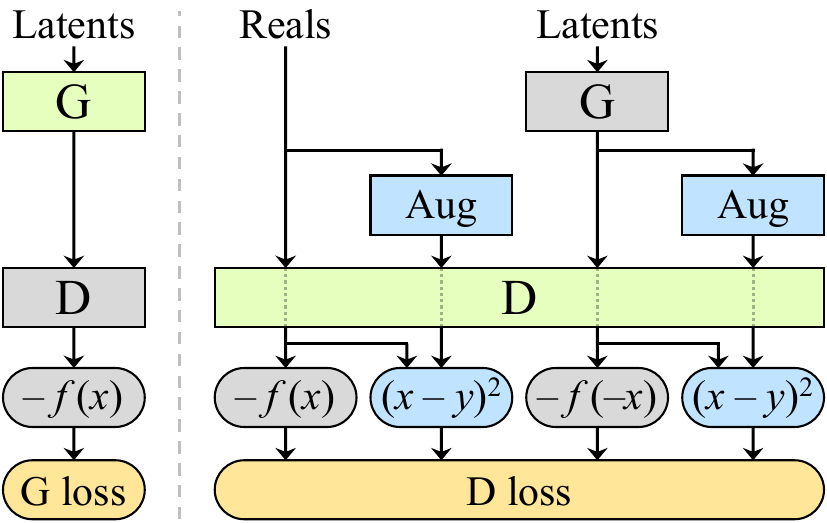}%
\vfill\makebox[\linewidth][c]{(a) bCR (previous work)}}%
\hspace{10mm}
\parbox[b][\vv]{\hh}{%
\includegraphics[width=1.0\linewidth]{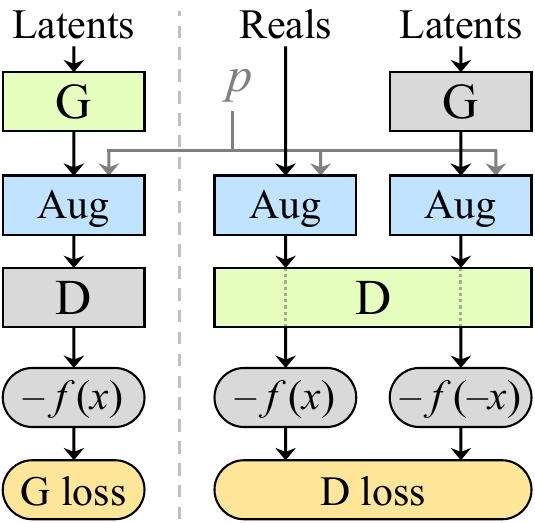}%
\vfill\makebox[\linewidth][c]{(b) ADA}}%
\caption{Flowcharts for (a) balanced consistency regularization (bCR)~\cite{zhao2021improved} and (b) the stochastic discriminator augmentations from~\cite{karras2020training}.}
\label{fig:ada}
\end{figure*}
}

\newcommand{\figFUNIT}{
\begin{figure*}[t]
\centering
\includegraphics[width=1.0\linewidth]{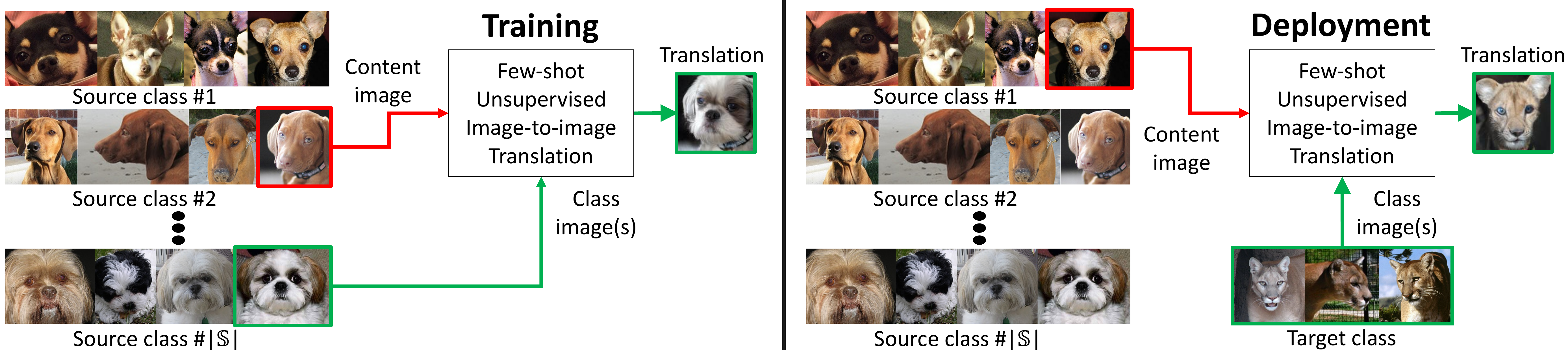} \\
\caption{The training and deployment scheme of FUNIT~\cite{Liu2019FewShotUI}. The purposed model aims to generate a translation of the input image that resembles images of the target class.}
\label{fig:funit}
\end{figure*}
}

\newcommand{\figLWg}{
\begin{figure*}[t]
\centering
\includegraphics[width=1.0\linewidth]{figs/lightweight_gan/g_overview.pdf} \\
\caption{The structure of the skip-layer excitation module and the Generator from \cite{liu2020towards}.}
\label{fig:lwg}
\end{figure*}
}

\newcommand{\figLWd}{
\begin{figure*}[t]
\centering
\includegraphics[width=1.0\linewidth]{figs/lightweight_gan/d_overview.pdf} \\
\caption{The structure and the forward flow of the Discriminator from \cite{liu2020towards}.}
\label{fig:lwd}
\end{figure*}
}

\newcommand{\figsingan}{
\begin{figure*}[t]
\centering
\includegraphics[width=0.7\linewidth]{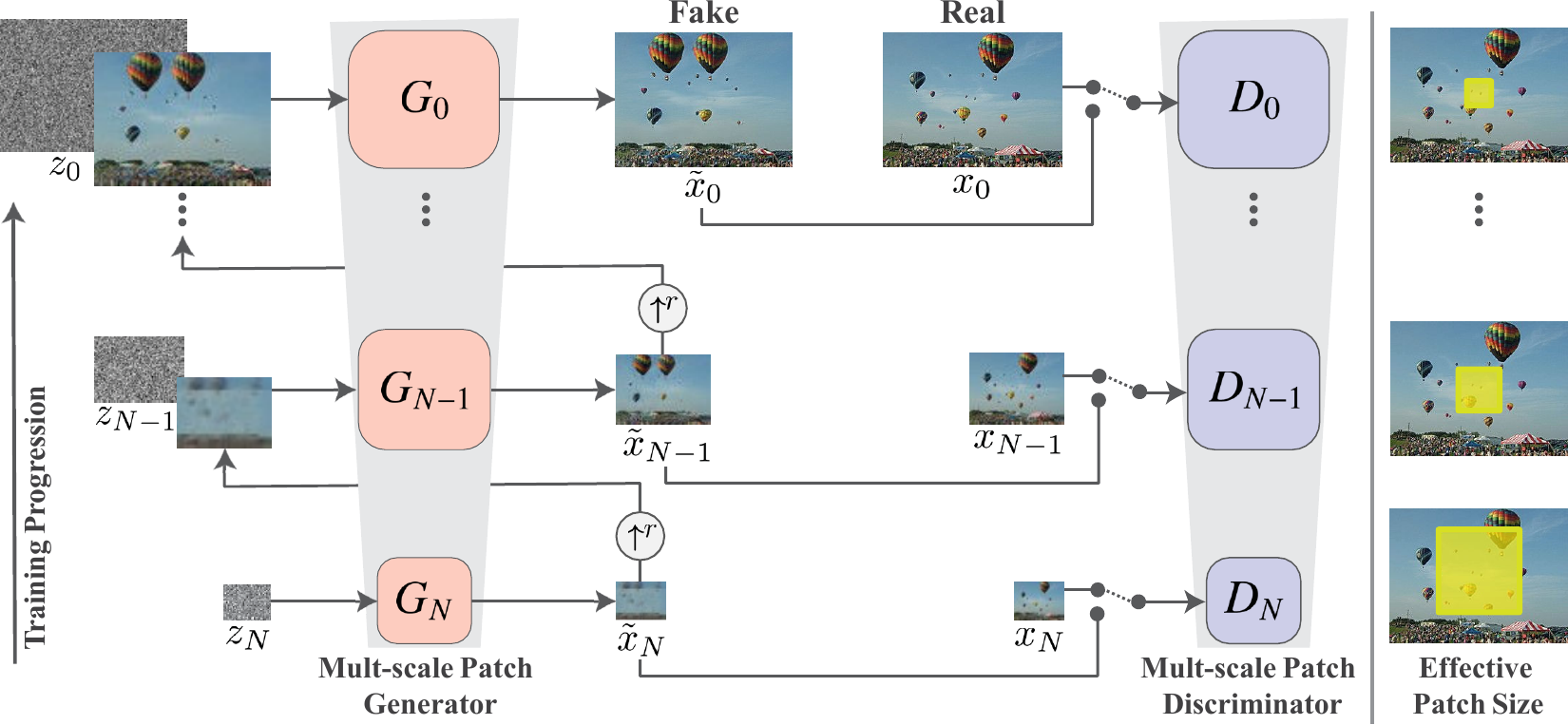} \\
\caption{The multi-scale pipeline of SinGAN~\cite{Shaham2019SinGANLA}.}
\label{fig:singan}
\end{figure*}
}

\newcommand{\figoneshotgan}{
\begin{figure*}[t]
\centering
\includegraphics[width=1.0\linewidth]{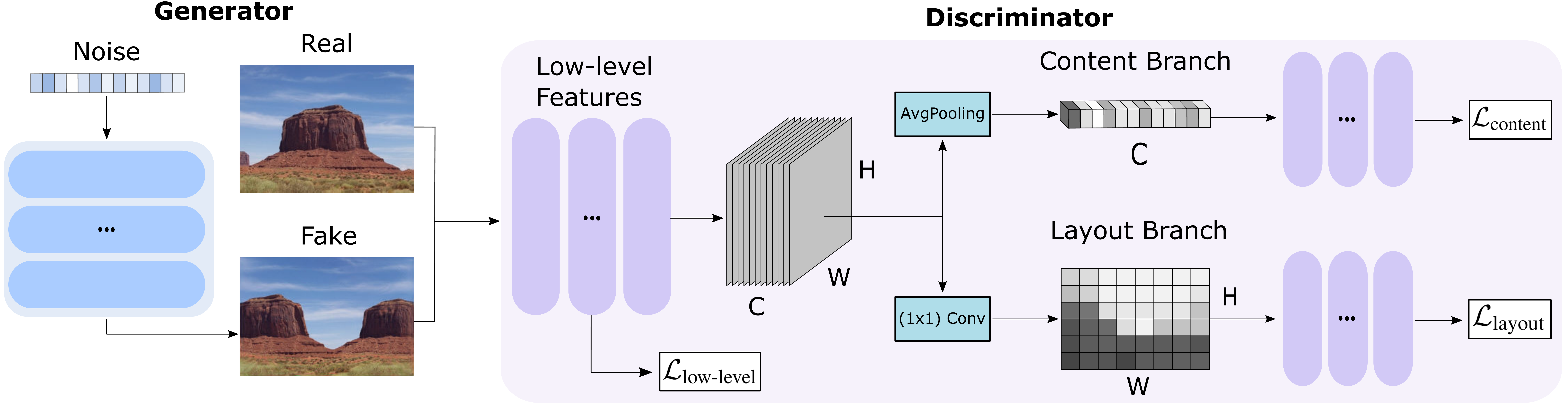} \\
\caption{The model overview of One-shot GAN~\cite{Sushko_2021_CVPR}.}
\label{fig:oneshotgan}
\end{figure*}
}

\newcommand{\figgpunit}{
\begin{figure*}[t]
\centering
\includegraphics[width=1.0\linewidth]{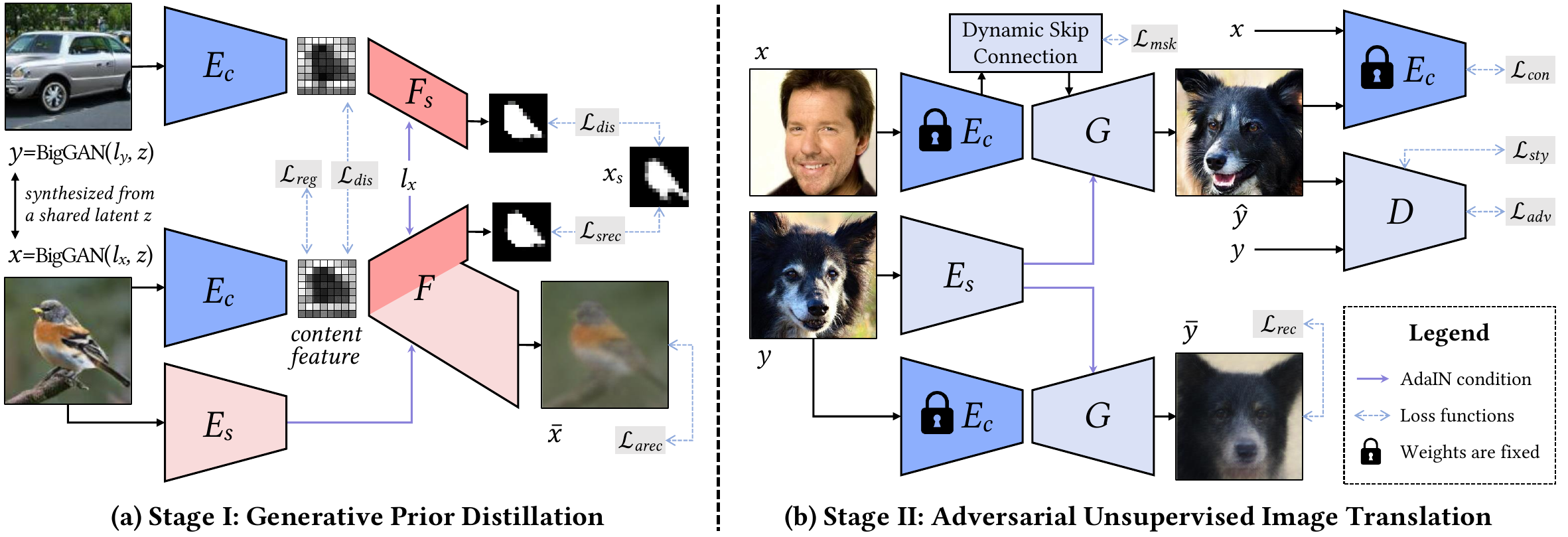} \\
\caption{Overview of the GP-UNIT~\cite{yang2022unsupervised}. (a) Stage I: Generative Prior Distillation. (b) Stage II: Adversarial Unsupervised Image Translation.}
\label{fig:gpunit}
\end{figure*}
}

\newcommand{\figbigdatasetgan}{
\begin{figure*}[t]
\centering
\includegraphics[width=1.0\linewidth]{figs/bigdatasetgan.pdf} \\
\caption{Architecture of BigDatasetGAN based on BigGAN.}
\label{fig:bigdatasetgan}
\end{figure*}
}

\newcommand{\figdatasetgan}{
\begin{figure*}[t]
\centering
\includegraphics[width=1.0\linewidth]{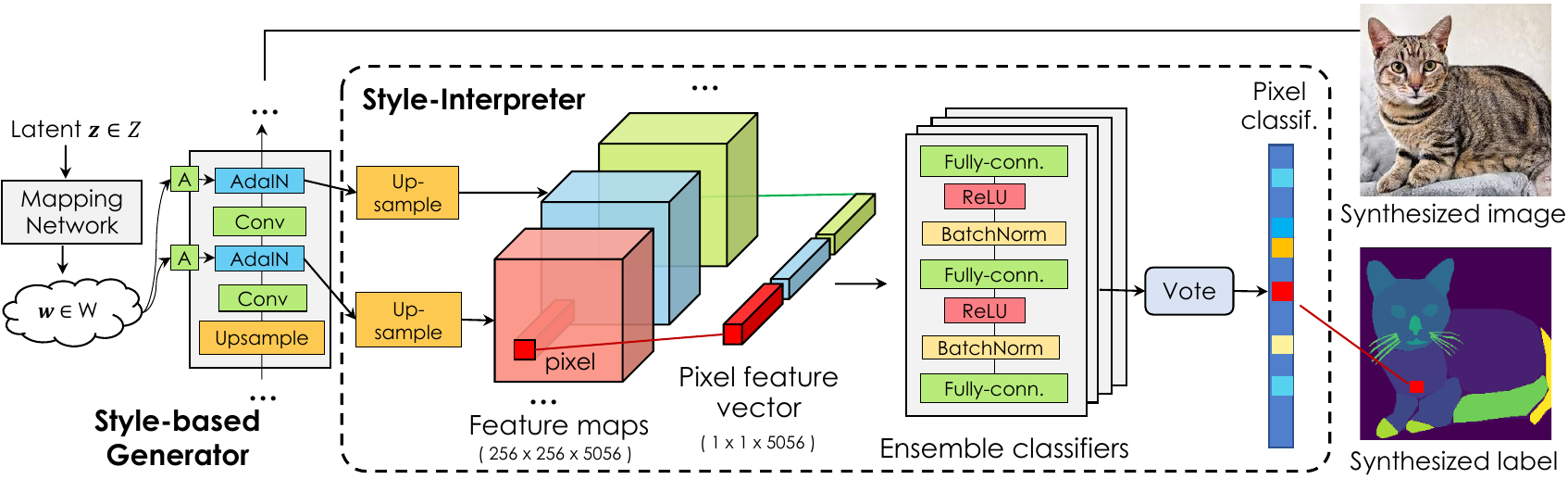} \\
\caption{Overall architecture of DatasetGAN~\cite{Zhang2021DatasetGANEL}. 
}
\label{fig:datasetgan}
\end{figure*}
}

\newcommand{\figganbasic}{
\begin{figure*}[t]
\centering
\includegraphics[width=0.8\linewidth]{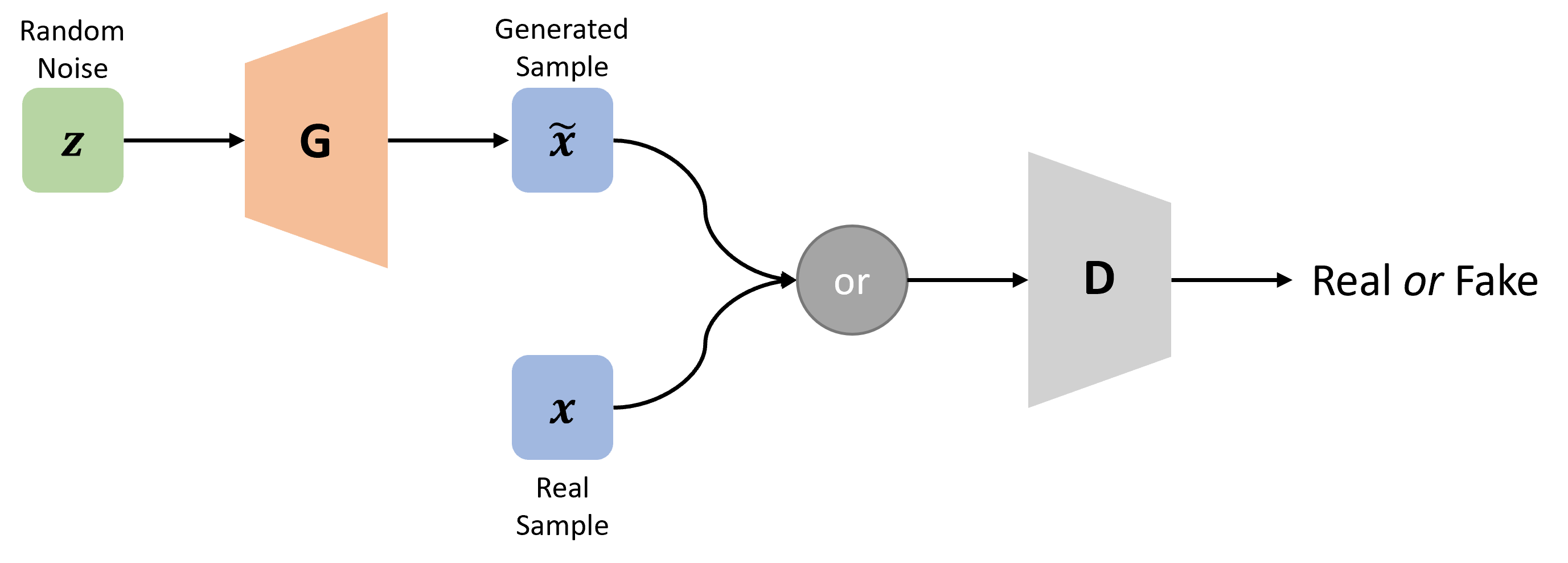} \\
\caption{The general structure of a generative adversarial network.}
\label{fig:ganbasic}
\end{figure*}
}


\newcommand{\figepsty}{
\begin{figure}[t]
    \centering
    \includegraphics[width=0.9\textwidth]{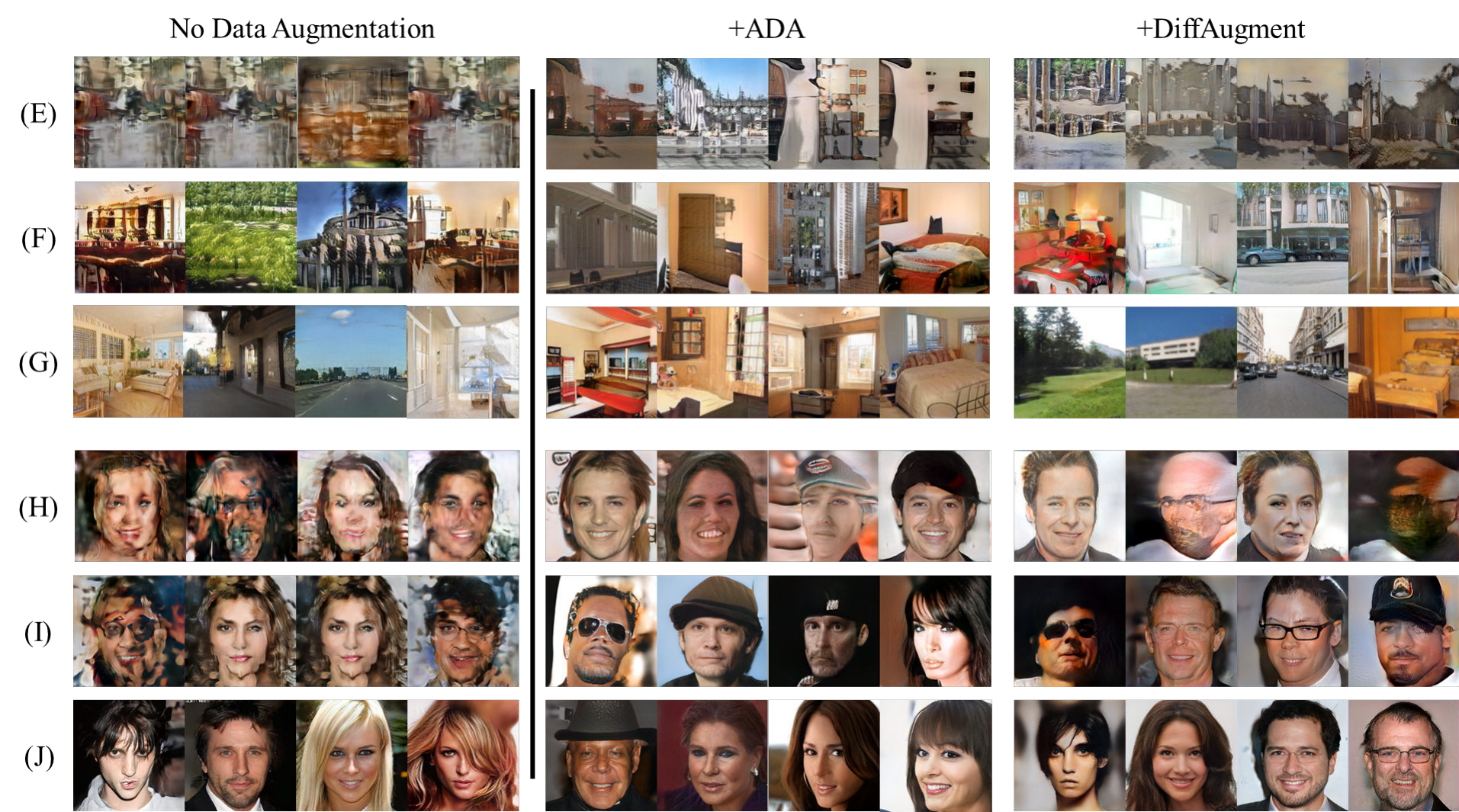}
    \caption{The sampled images from StyleGAN2 with models trained under different settings.}
    \label{fig:epsty}
\end{figure}
}

\newcommand{\figepspade}{
\begin{figure}[t]
    \centering
    \includegraphics[width=0.9\textwidth]{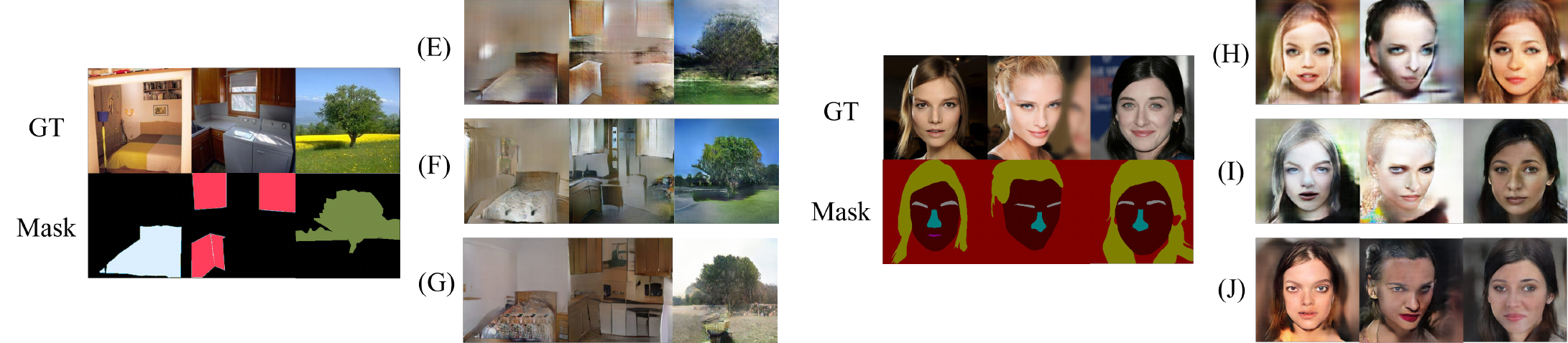}
    \caption{The sampled images from SPADE with models trained under different settings.}
    \label{fig:epspade}
\end{figure}
}

\newcommand{\figepsean}{
\begin{figure}[t]
    \centering
    \includegraphics[width=0.9\textwidth]{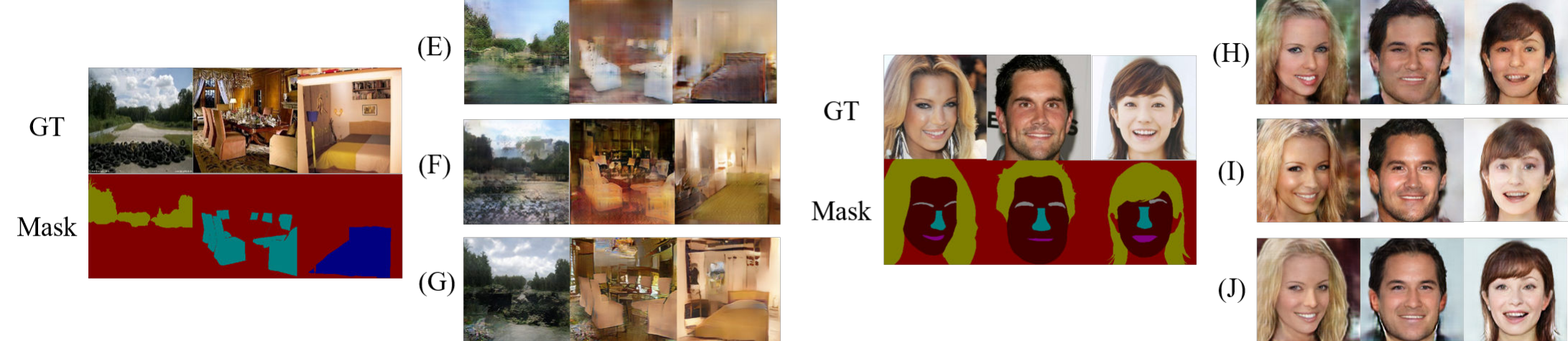}
    \caption{The sampled images from SEAN with models trained under different settings.}
    \label{fig:epssean}
\end{figure}
}

\newcommand{\figepfunit}{
\begin{figure}[t]
    \centering
    \includegraphics[width=0.9\textwidth]{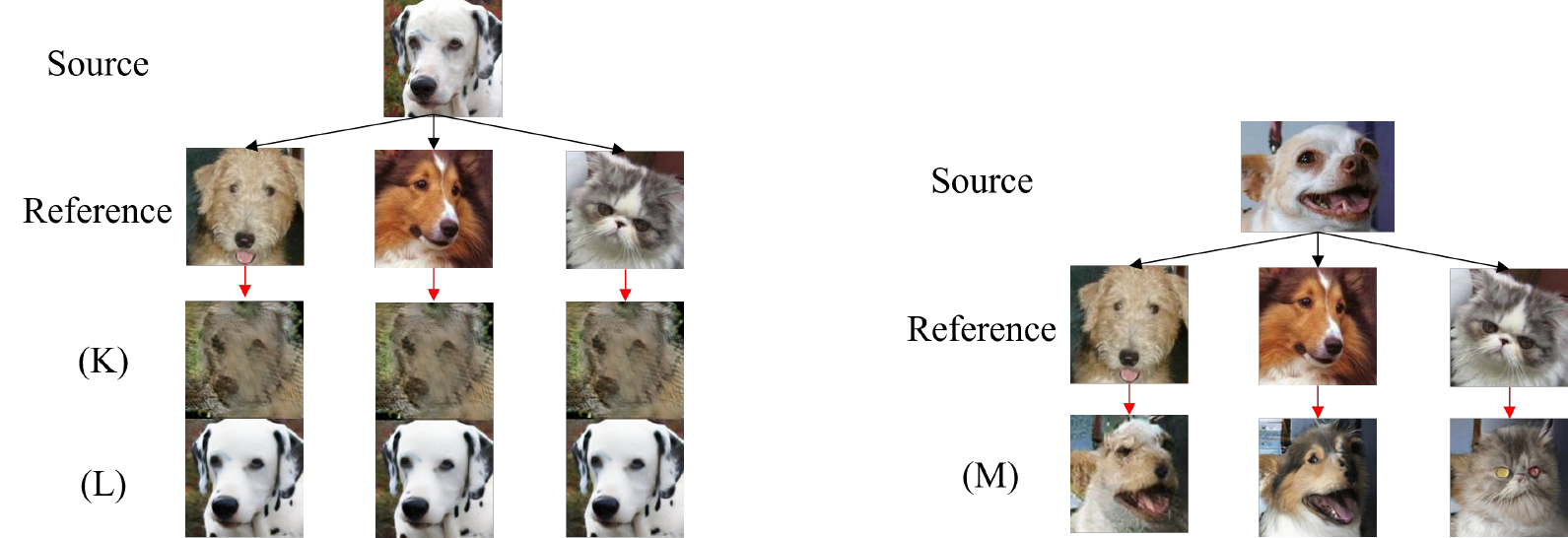}
    \caption{The sampled images from FUNIT with models trained under different settings.}
    \label{fig:epfunit}
\end{figure}
}

\newcommand{\figepbiggan}{
\begin{figure}[t]
    \centering
    \includegraphics[width=1.0\textwidth]{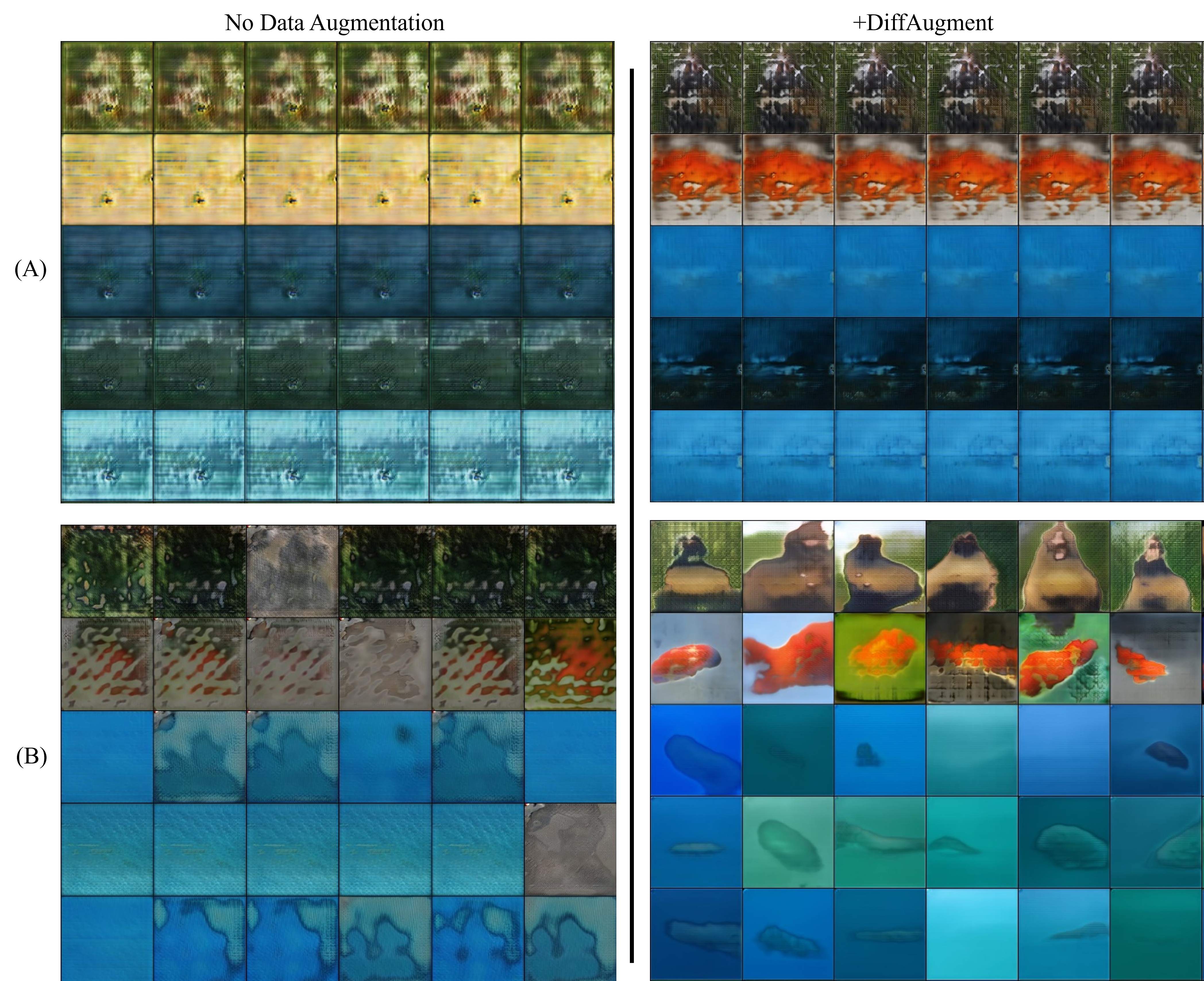}
    \caption{The sampled images from BigGAN with models trained under different settings, where the rows stand for the first five classes of ImageNet---\textit{Tench}, \textit{Goldfish}, \textit{Great white shark},\textit{Tiger shark}, and \textit{Hammerhead shark}, respectively. }
    \label{fig:epbiggan}
\end{figure}
}

\newcommand{\figepbigganCeleb}{
\begin{figure}[t]
    \centering
    \includegraphics[width=0.9\textwidth]{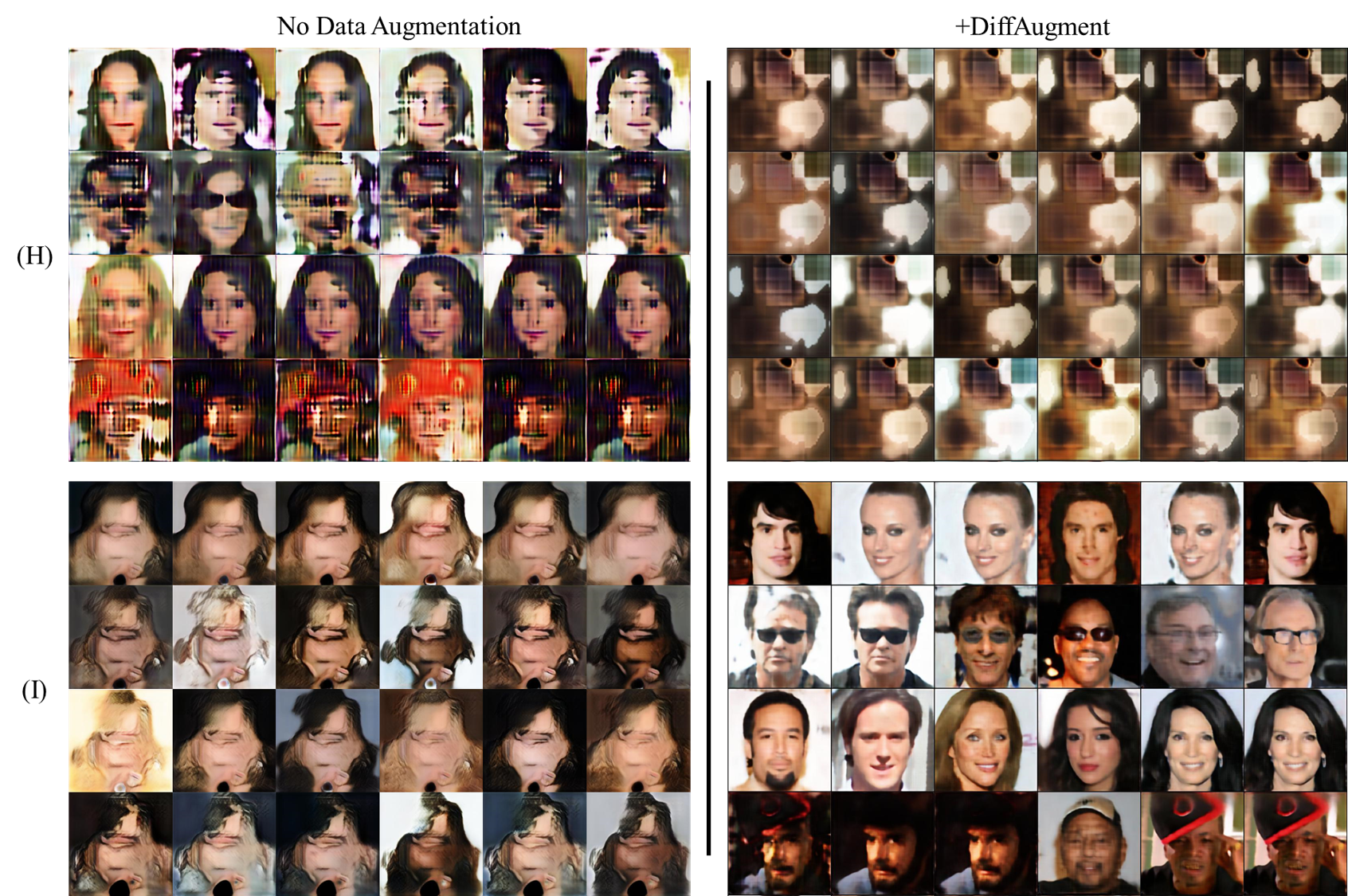}
    \caption{The sampled images from BigGAN with models trained under different settings for the seven classes. }
    \label{fig:epbiggan_celeb}
\end{figure}
}

\newcommand{\figepstarganceleba}{
\begin{figure}[t]
    \centering
    \includegraphics[width=0.9\textwidth]{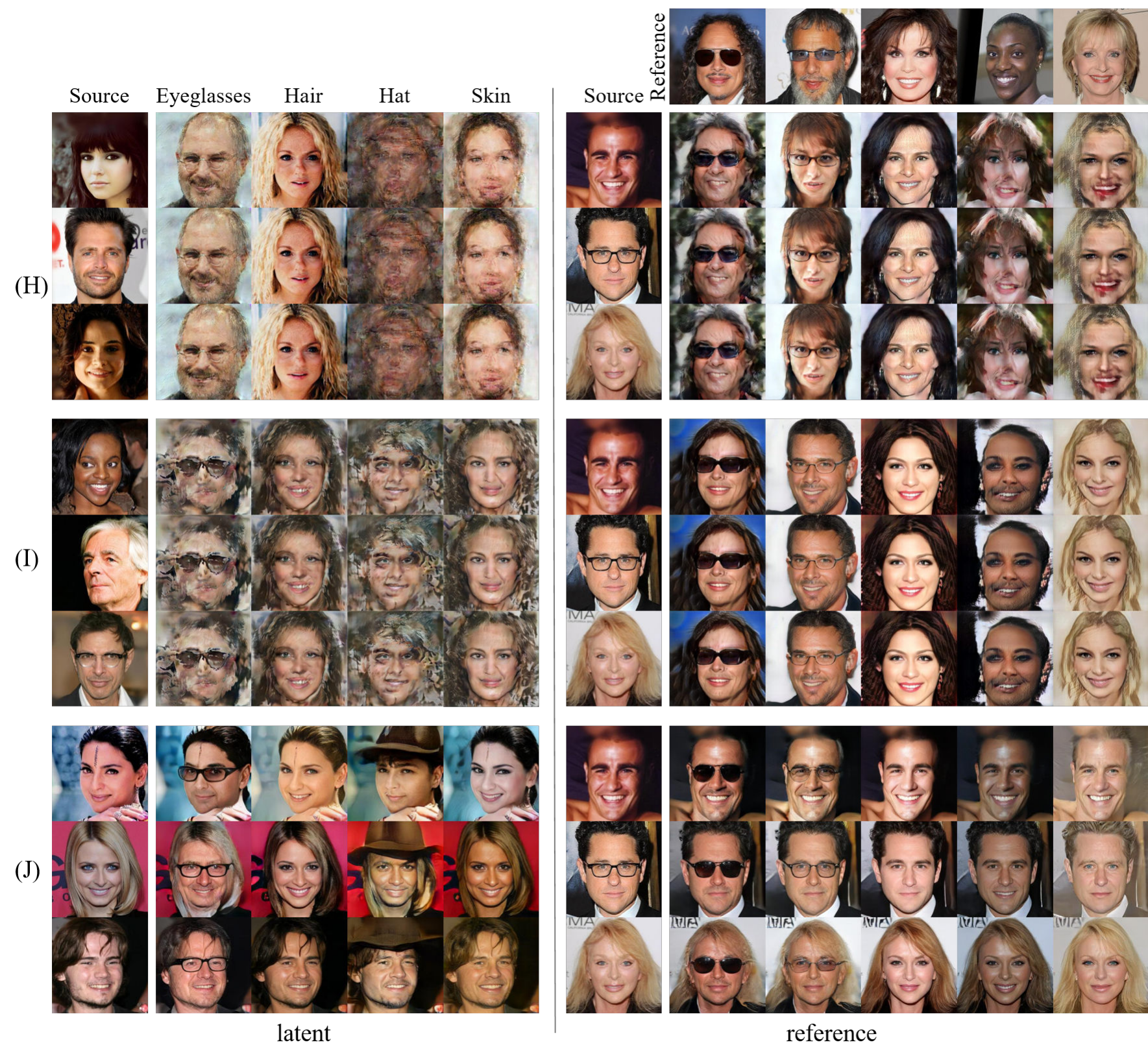}
    \caption{Images sampled from StarGAN v2 of CelebAMask-HQ dataset with models trained under different settings.}
    \label{fig:epsstargan_celeb}
\end{figure}
}

\newcommand{\figepstarganafhq}{
\begin{figure}[h]
    \centering
    \includegraphics[width=0.5\textwidth]{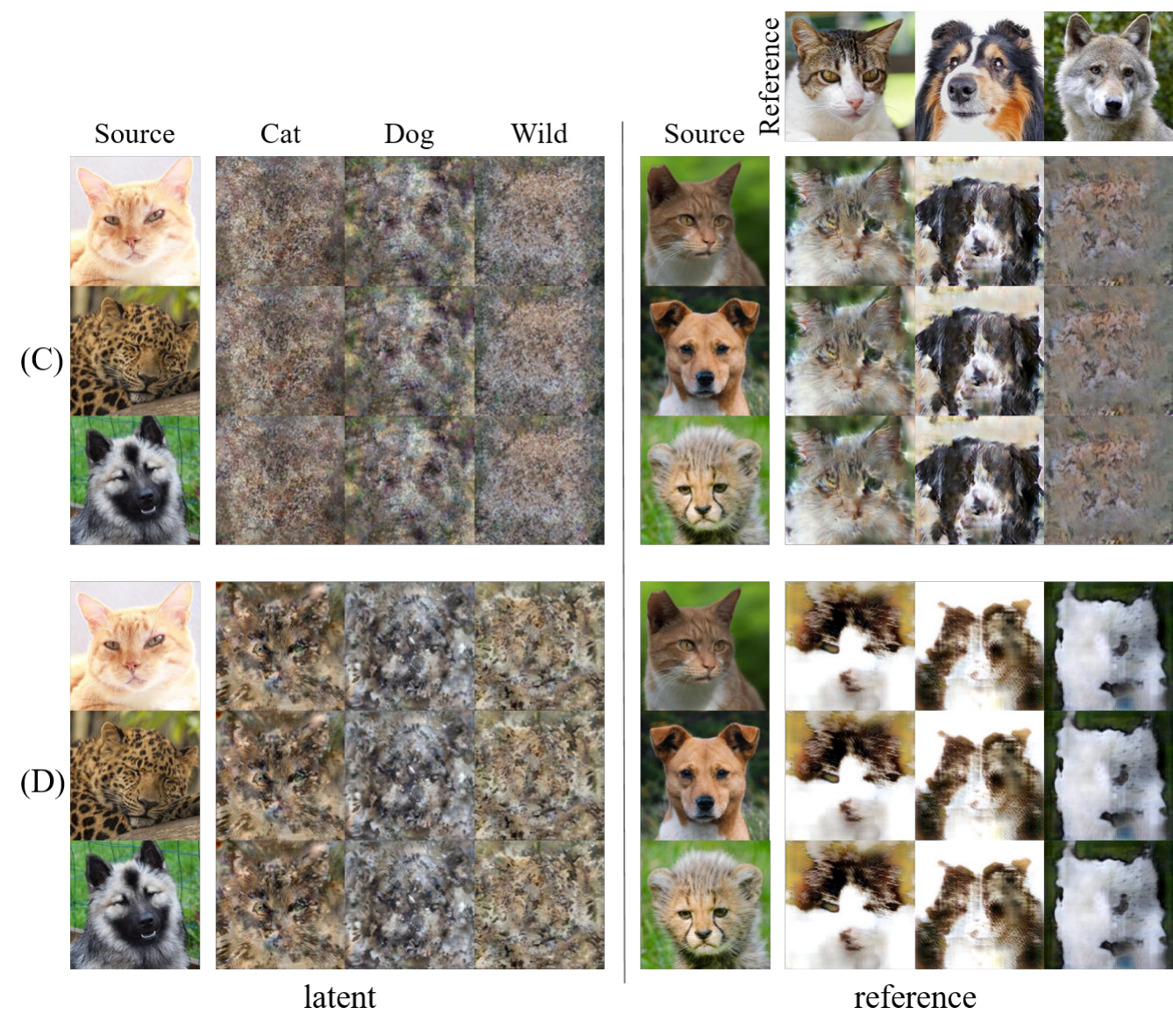}
    \caption{Images sampled from StarGAN v2 of AFHQ dataset with models trained under different settings.}
    \label{fig:epsstargan_afhq}
\end{figure}
}

\newcommand{\figepstargananimals}{
\begin{figure}[h]
    \centering
    \includegraphics[width=0.9\textwidth]{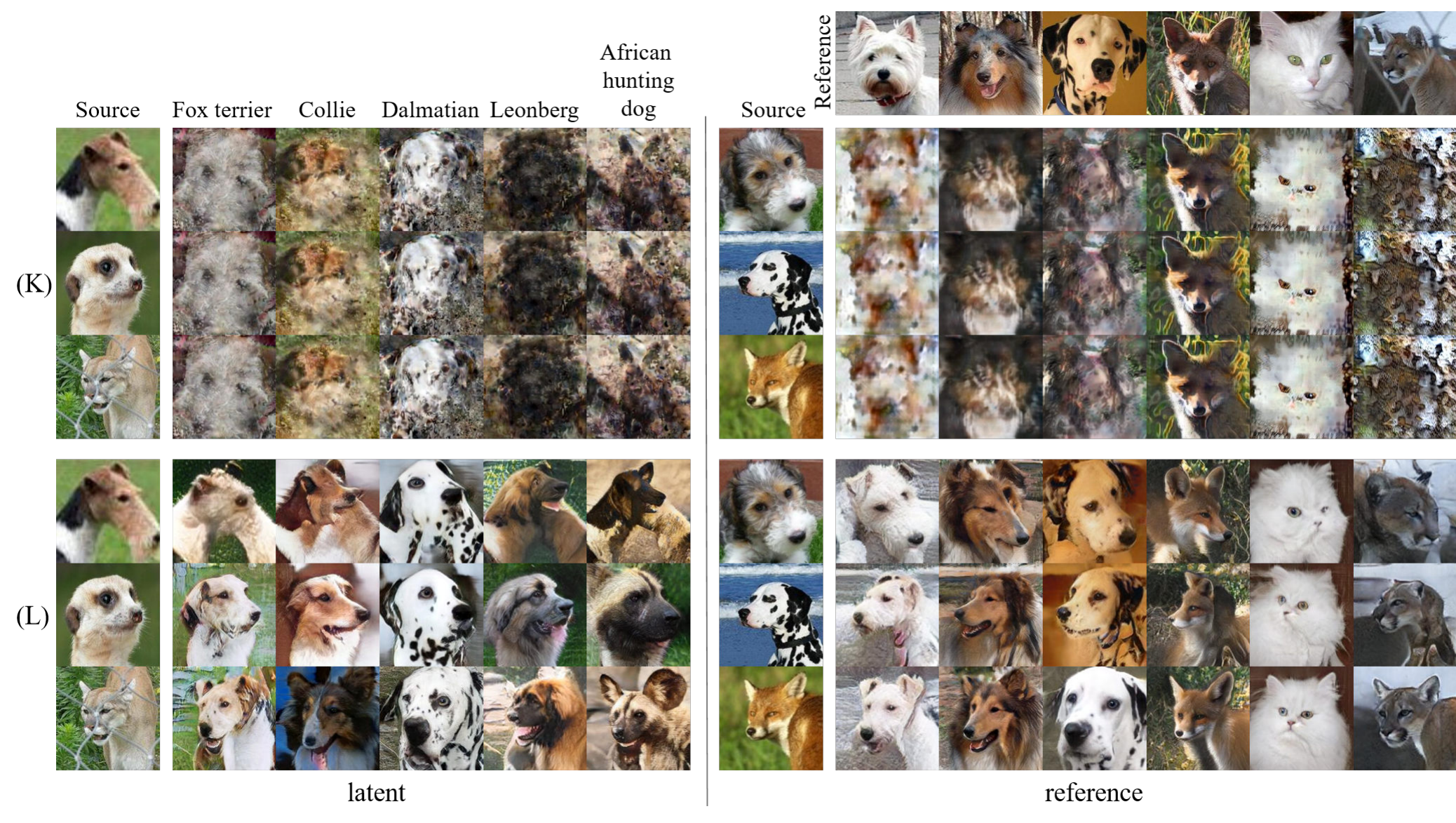}
    \caption{Images sampled from StarGAN v2 of Animal Faces dataset with models trained under different settings.}
    \label{fig:epsstargan_animals}
\end{figure}
}

\title[GANs with Limited Data: A Survey with Benchmarking]{Generative Adversarial Networks with Limited Data: A Survey and Benchmarking}


\author[1,3]{\fnm{Omar} \sur{De Mitri}}\email{omar.de.mitri@ipa.fraunhofer.de}
\equalcont{These authors contributed equally to this work.}

\author[2]{\fnm{Ruyu} \sur{Wang}}\email{emmawang9211@gmail.com}
\equalcont{These authors contributed equally to this work.}

\author[1,2]{\fnm{Marco F.} \sur{Huber}}\email{marco.huber@ieee.org}

\affil[1]{\orgname{Fraunhofer IPA}, \orgaddress{\city{Stuttgart}, \country{Germany}}}

\affil[2]{\orgdiv{Institute of Industrial Manufacturing and Management (IFF)},\orgname{University of Stuttgart}, \orgaddress{\city{Stuttgart}, \country{Germany}}}

\affil[3]{\orgdiv{Dept. of Innovation Engineering}, \orgname{University of Salento}, \orgaddress{\city{Lecce}, \country{Italy}}}


\abstract{Generative Adversarial Networks (GANs) have shown impressive results in various image synthesis tasks. Vast studies have demonstrated that GANs are more powerful in feature and expression learning compared to other generative models and their latent space encodes rich semantic information. However, the tremendous performance of GANs heavily relies on the access to large-scale training data and deteriorates rapidly when the amount of data is limited. This paper aims to provide an overview of GANs, its variants and applications in various vision tasks, focusing on addressing the limited data issue. We analyze state-of-the-art GANs in limited data regime with designed experiments, along with presenting various methods attempt to tackle this problem from different perspectives. Finally, we further elaborate on remaining challenges and trends for future research.
}




\maketitle
\section{Introduction}\label{sec1}
Since their introduction, GANs have 
attracted more and more attention in the field of deep generative models.
Their success is mainly due to their ability to learn the features of a training dataset and to be able to generate images of increasing quality.
Numerous architectures have been developed over the years in many areas of computer vision. Examples can be found for image synthesis, super-resolution, in-painting, etc. 
Although most of these architectures are capable of expressing the best results when trained with very large datasets, some problems may arise when they are trained with a smaller amount of data.

Typical problems can be a lack of diversity in the generated images, caused by `mode collapse' or `overfitting' of the network.
In some cases, it is not uncommon to find network instability with a consequent degrading of the image quality generated by the network.
These problems can be critical when trying to transfer this technology to real-world applications. In real-world scenarios, it is difficult to access large amounts of data for training purposes. 
The acquisition of large quantities of data for training the network is costly in terms of time, human, and budget resources. In addition, in certain contexts such as manufacturing or healthcare, a low occurrence of anomalous events can be observed with consequent difficulty in acquiring large amounts of data.
For this reason, the development of GAN architectures capable of generating good quality images with an ever smaller amount of data is essential. 
The literature offers numerous reviews about GAN theory and applications \cite{gui2021review, park2021review, wang2020state, alotaibi2020deep}. However, we are not aware of any work analyzing these architectures under a limited data regime.

In this paper, we 
collect the most recent GAN architectures used in computer vision and evaluate their performance under different conditions of data scarcity. The questions we wanted to answer are: (1) How do GAN architectures perform in the presence of limited datasets? (2) How does their performance degrade under these conditions? (3) What type of architecture performs best under these conditions?
The contributions of this paper are as follows:
\begin{itemize}
    \item Provide an introduction to GANs and the problems encountered in their training with limited datasets.
    \item Present the state of the art of GAN architectures, also analyzing the various strategies used to deal with the condition of scarcity.
    \item Analyze and compare the performance of some benchmark GAN architectures in different data scarcity scenarios.
\end{itemize}


\section{Generative Adversarial Networks}\label{sec2}

The generative adversarial network has been one of the significant recent developments in 
deep generative models. Unlike traditional generative models (e.g., Gaussian mixture models (GMM)\citep{rasmussen1999infinite}) which cannot perform well on complex distributions, deep generative models utilize techniques such as deep neural networks and stochastic backpropagation to learn variational distributions from large-scale datasets \citep{maaloe2016auxiliary, oussidi2018deep}. The vanilla GAN was proposed in 2014 by Goodfellow et al. \cite{goodfellow2014gan}, where the fundamental aspect of it is a min-max two-player zero-sum game. The general structure of a GAN consists of two competing subnetworks---a generator $G$ and a discriminator $D$ as illustrated in \autoref{fig:ganbasic}. During the training phase, the goal of the generator $G$ is to deceive the discriminator $D$ with the samples it generates from randomly sampled noise. Meanwhile, the discriminator $D$ is tasked to distinguish between real samples from the training set and the fake ones generated by the generator $G$. The main aim of the whole training process is to achieve the Nash equilibrium \citep{ratliff2016characterization}. The objective function of the vanilla GAN is formulated as
\begin{equation}
\min_{G} \max_{D} \mathcal{V}(G,D)={E_{x\sim P_\mathrm{data}}}[\log D(x)] + {E_{z\sim P_{z}}}[\log(1-D(G(z)))]~,
\end{equation}
where $P_\mathrm{data}$ is the true data distribution and $P_z$ is the noise distribution.

\figganbasic

Despite the superior performance of GANs in many image synthesis tasks against other generative methods, GAN training is notorious for its instability. Goodfellow et al. \cite{goodfellow2014gan} had provided the theoretical proof shown the existence of unique solutions, where the generator $G$ is optimal when $P_{z} = P_\mathrm{data}$ and the discriminator $D$ is predicting a classification score of 0.5 for all samples drawn from $x$. However in practice, GAN training is still challenging and unstable for several reasons such as: 
\begin{itemize}
\item Difficulties in convergence  \citep{radford2015unsupervised} for both the generator and the discriminator.
\item Mode collapsing \citep{salimans2016improved}, where the network produces a sole output despite various inputs being given. 
\item Zero gradient \citep{arjovsky2017towards}, where the discriminator loss converges quickly to zero and thus, provides the generator with no reliable cue for gradient updates.
\end{itemize}
Several researchers have proposed variety of solutions that addressed these issues from different aspects: improved formulations of the objective functions, the model structure, the regularization methods.

\textbf{Objective Functions.} Nowozin et al. \cite{nowozin2016f} showed that GAN training may be generalized to minimize an estimate of f-divergences such as KL-divergence and proposed an alternative objective to replace the vanilla one which was easily saturated at the beginning of the training due to the weak gradients. Arjovsky et al. \cite{arjovsky2017wasserstein} proposed to prevent gradient vanishing by a novel cost function deriving from an approximation of the Wasserstein distance. The main idea of the proposed Wasserstein GAN (WGAN)  relied on the discriminator being a k-Lipschitz continuous function, which in practice can be  implemented by simply clipping the parameters of the discriminator. However, 
a later work \citep{gulrajani2017improved} showed that weight clipping reduced the capacity of the model to learn more complex functions. Gulrajani et al. thus proposed WGAN-GP to penalize the norm of discriminator gradients with respect to data samples during training instead of simply clipping the weights.

\textbf{Model Structures.}  Radford et al. \cite{radford2015unsupervised} introduced the Deep Convolutional GAN (DCGAN) architecture, which led to major improvements in stabilizing GAN training. It was the first work to combine a GAN with a convolutional neural network (CNN) instead of the multilayer perceptron used in the vanilla GAN, allowing the model to learn spatial relationships for high-quality image generation. Moreover, several choices of design such as using batch normalization (BN) and ReLU activation functions as well as removing fully connected hidden layers were recommended by the authors for increasing model stability and performance. Later, Self-Attention GAN (SAGAN) \citep{zhang2019self} was proposed to incorporate a self-attention mechanism as an aid to convolutions for modeling long-range, multi-level dependencies across image regions. Progressive Growing GAN (PGGAN)\citep{Karras2017ProgressiveGO} further improved the performance of GANs in high-resolution generation by introducing a training scheme that added new blocks of layers progressively to both the generator and the discriminator during training. Moreover, the proposed progressive training not only stabilized the learning process but also reduced the training time. Beside training unconditionally, several works have proposed to integrate conditional signal into GANs in order to have control over the generated images. Conditional GAN (cGAN)\citep{mirza2014conditional} was the first work that combined a random noise $z$ and a conditional variable $c$ into a joint hidden representation of real data $x$ (i.e. $G(z,c)$ instead of $G(z)$) and performed conditional discrimination in discriminators, which provided better representation than DCGAN in generating various data. The explicitly usage of the condition variable $c$ turned the GAN training into a supervised manner. Chen et al. \cite{chen2016infogan} thus proposed Information GAN (InfoGAN), which learned to disentangle the incompressible noise vector $G(z)$ and latent variable $c$ in an unsupervised manner. The conditional latent variable $c$ of InfoGAN was no longer given but to be discovered through training. By maximizing the mutual information between the generator's output $G(z,c)$ and latent code $c$, InfoGAN was able to discover the meaningful features of real data distribution while remaining unsupervised.  

\textbf{Regularization Methods.}
In GAN models, regularization methods like weight penalization have been extensively used to prevent the mode collapse problem. Brock et al. \cite{brock2016neural} proposed a novel Orthogonal Regularization (OR) as a weight penalty for the objective function to replace $L_2$ norm which harmed the performance. Miyato et al. purposed Spectral Normalization (SN)\cite{miyato2018spectral} to normalize the weight matrices and did not use additional losses, which was later commonly employed in the literature. 

Extensive effort has been made to improve and stabilize the GAN training process, however, most of the works were conducted on large-scale datasets with balanced and abundant data for the model to learn. When training on datasets with only a handful of samples, GANs still suffer from the aforementioned problems---difficulties in convergence, mode collapse, and zero gradient. Moreover, the nature of limited datasets brings new challenges to overcome:
\begin{itemize}
\item Overfitting \citep{karras2020training}, where the network can only reproduce samples from the training set.
\item Lack of diversity due to learning from sparse or imbalanced number of data points. 
\end{itemize}

In this survey, we aim to evaluate models and methods that were originally trained and evaluated on large-scale datasets on several hand-crafted small datasets and provide insights on the limitations of existing methods, the open challenges, and potential directions for future research.

\section{State-of-the-art Application Models}\label{sec3}
Generative Adversarial Networks have shown remarkable performance across various domains, enabling the synthesis of high-quality, photorealistic images, seamless style transformations, and robust image-to-image translation. This section comprehensively reviews state-of-the-art application models in GANs, categorizing them based on their primary objectives and methodologies.
The section is divided into two main subsections. The first subsection, Image Synthesis, covers GAN models for generating high-resolution and realistic images. The second subsection, Image-to-Image Translation, explores GAN models that learn mappings between different image domains, facilitating tasks such as style transfer, semantic segmentation, and domain adaptation. 
\subsection{Image Synthesis}\label{subsec2}
Image synthesis is one of the most exciting applications for generative networks allowing the generation of new instances of high-resolution, realistic and colourful pictures.
Most of the architectures released can be distinguished into three categories: unconditional, conditional, and semantic.
In the unconditional version, architectures synthesize the image based on the training distribution without any condition/information about the image to be generated. 
In the conditional version, on the other hand, information about the image class is integrated into the architecture by conditioning the network to produce the image to be output.
Finally, semantic image synthesis is a variant of conditional GAN in which the network is conditioned through a semantic layout. 

\subsubsection{StyleGAN}
StyleGAN is a family of architectures for high-resolution unconditional image synthesis.
In StyleGAN, Karras et al. \cite{Karras_2019_CVPR} proposed a new structure of a generator consisting of two blocks: a mapping network \emph{f} and a synthesis network \emph{g}. 
The mapping network aimed to learn the different styles from a learned distribution, while the synthesis network aimed to generate new images based on a style collection.
As shown in \autoref{fig:stylegan1_arch}, the mapping network \emph{f}, comprising eight MLP layers, has as input the latent vector $z \in \mathcal{Z}$ and as output the intermediate latent space $\mathcal{W}$. The learned affine transformation in $w \in \mathcal{W} $ specialized the latent vector $w$ to the data styles $y=(y_{s},y_{b})$. They were used to feed and control each level of the synthesis network via adaptive instance normalization (AdaIN).
The AdaIN operation is built in the synthesis network \emph{g} and followed the convolutional layers. It first normalizes each channel to zero mean and unit variance and then applies scales $y_{s}$ and biases $y_{b}$ based on the style.

Although the images generated by StyleGAN achieved a high level of quality, they often produced artifacts similar to water drops. The problem, analyzed by the authors in StyleGAN2 \cite{Karras_2020_CVPR}, was attributed to the way the average and normalization operations were carried out in the AdaIN layer.
Therefore, the internal structure of the style block was modified, moving out the operation of adding noise and biases outside the style block and integrating the normalization operation into a convolution layer.
The revised synthesis network is represented on \autoref{fig:stylegan2_arch}. 

In StyleGAN3 \cite{Karras.2021}, the researchers observed an unintentional positional references of features in the intermediate layers of the StyleGAN2. In fact, it was observed that the coarse network features controlled the presence of the finer ones but do not manage the position dependency, which was fixed in terms of pixel coordinates.
For this purpose, the internal architecture was redesigned to eliminate all sources generating positional references and to make the network \emph{equivariant}.
An operation $f$ like convolution, upsampling, ReLU, is called equivariant with respect to a spatial transformation $t$ of the 2D plane if it commutes with it in the continuous domain: \begin{math}t \circ f = f \circ t\end{math}.
The modified architecture StyleGAN3  exhibited a more natural transformation hierarchy, where the exact sub-pixel position of each feature was exclusively inherited from the underlying coarse features and it was more indicated for video and animation applications.

\figStyleGANs

\subsubsection{BigGAN}
BigGAN \cite{Brock.25.02.2019} is an architecture for class-conditional image synthesis based on 
SA-GAN \cite{zhang2019self} (\autoref{fig:BigGAN_arch}).
The class information is provided to the generator $G$ using a class-conditional BatchNorm and to the discriminator $D$ using projection. 
The input latent vector $z$ is split along its channel into equal chunks, and each chunk is then concatenated to a shared class embedding and passed to the corresponding residual block.

The class-conditional BatchNorm, called also \textit{conditional instance normalization} \cite{Dumoulin.2016} allows a layer's activation $x$ to be transformed to a normalized activation $a_\mathrm{cin}$ specifying the painting style $s$.
During their experiments, the authors observed improvements in performance by increasing the batch size and allowing the network to have more data variance per batch. However, some training instability problems were observed, mainly at a high number of iterations, which can be solved by using early stopping.

\figsBigGAN

\subsubsection{VQ-GAN}
Unlike StyleGAN and BigGAN, Esser et al. propose a novel approach that differs from traditional methods like StyleGAN and BigGAN, which primarily rely on convolutional neural networks (CNNs). Their method---termed VQ-GAN \cite{Esser2021TamingTF}---integrates transformer architecture to better understand complex relationships among inputs. In contrast to CNNs, which have a built-in preference for local interactions, transformers lack this inductive bias, allowing them to capture complex relationships that extend beyond local contexts. However, this flexibility comes with the challenge of learning all potential interactions, which can be computationally overwhelming for long sequences such as high-resolution images.

To address this, the authors suggest combining the strengths of both CNNs and transformers as depicted in \autoref{fig:VQGAN}: utilizing CNNs to develop a codebook of rich visual parts efficiently and then employing transformers to model their global compositions. This combination allows for long-range interactions within these compositions, necessitating a more expressive transformer architecture to represent the distributions of the visual components. Additionally, the authors also employ an adversarial approach to ensure the local parts encoded by the convolutional method capture perceptually important structures, reducing the need for modeling low-level statistics with transformers. This novel network design thus allows VQ-GAN to generate high-resolution images efficiently.
\figVQGAN

\subsubsection{SemanticStyleGAN}
With SemanticStyleGAN~\cite{Shi.2022}, Shi, Yang et al. developed an architecture for semantic image synthesis with separate modeling of all image components. 
Based on StyleGAN2 \cite{Karras_2020_CVPR}, the authors extended the space $\mathcal{W}^{+}$ into different semantic areas $\mathcal{W}^{K}$,  
where each local latent code $w^k \in \mathcal{W}^{k}$ was decomposed to control shape ($w_{s}^{k}$) and texture ($w_{t}^{t}$) of every semantic area $k \in K$. 
Each latent code $w^k$, together with position encoding information, was then used for a local generator $g_k$ to output a features map $f_k$ and a pseudo-depth map $d_k$.
\autoref{fig:SemanticStyleGAN} shows an overview of the architecture.
During the training, style mixing was performed in order to encourage interaction between different local parts, shapes, and textures.
The features obtained were then assembled in a fusion step.
At first, the pseudo-depth masks were used for generating a coarse segmentation map \begin{math} m \in \mathbb{R}^{K \times H^{c} \times W^{c}}\end{math}.
Then, a feature map $f$ was obtained aggregating the $K$ element-wise multiplication between pixels of the $k$-th class of $m$ ($m_k$) and feature maps $f_k$.
In the end, the render net $R$, similar to the StyleGAN2 generator, has two tasks: generate an output image based on the input feature map $f$ and refine the coarse segmentation mask $m$ into a final mask having the same size as the output image.

The two outputs are then used for the dual branch discriminator $D(x,y)$ with mainly two convolution branches: one for the segmentation mask and the other one for the synthesized image.
\figSemanticStyleGAN

\subsubsection{SPADE}
Park et al. released SPADE \cite{park2019semantic}, an architecture to perform semantical image synthesis. The work proposed a spatial-adaptive normalization layer (SPADE) that uses segmentation masks to modulate layer activation. This layer is a generalization of the BatchNorm and AdaIN normalization layers (\autoref{fig:spade_spatial_ada_morm}), with the difference of using a semantic input instead of image and having spatially variant parameters.
The goal is to learn a mapping function that converted an input semantic mask into a realistic image. 

The SPADE generator is based on ResNet blocks with upsampling layers. The SPADE residual block (ResBlk) is shown in \autoref{fig:spade_resblk}. Each block is integrated with two normalization layers that receive segmentation maps as input, allowing the activation functions of the different layers to be modulated. The SPADE activation function integrates standard normalization with spatially adaptive modulation driven by semantic maps. Feature maps are initially normalized, followed by the application of a spatially adaptive affine transformation. The parameters for this transformation are derived through convolutional layers operating on the semantic maps and passed to the layers at different scales.
Finally, the discriminator is based on the one used in Pix2pixHD \cite{pix2pixHD} but using the hinge loss term instead of the least squared loss term.

\subsubsection{SEAN}
Zhu et al. extendes what was proposed in SPADE \cite{park2019semantic} presenting a new type of normalization called \emph{Semantic Region-Adaptive Normalization} (SEAN) \cite{zhu2020sean}. SEAN normalization receives as input a segmentation map $M$ and a set of per-region style codes $ST$. The last one is generated by a style encoder that receives images and segmentation maps as input and returns a style matrix $ST$ as output.
The segmentation map and the style codes are then used to compute a style map, where each pixel is associated with a style vector (\autoref{fig:sean_norm}). The style map is then used to compute two modulation parameters $\beta$ and $\gamma$ of the activation layers similar to what is applied in SPADE.

Similar to the parent architecture \cite{park2019semantic}, the SEAN generator is also composed of a series of SEAN ResNet blocks with upsampling layers. An overview of the architecture is shown in \autoref{fig:sean_pipeline}. Revised from the structure of SPADE, each SEAN ResBlk contains three SEAN normalization blocks which, after receiving ST style codes and M segmentation maps as input, allow the modulation of scale and bias in the three convolutional layers.
In addition, similar to StyleGAN \cite{Karras_2019_CVPR, Karras_2020_CVPR}, noise is added after each normalization block with the goal of improving the quality of the synthesized image.

\figSpadeSean

\subsection{Image-to-Image Translation}
\label{sec:i2i-gan}
Image-to-image translation using GANs has made great progress in both supervised and unsupervised learning research in the past few years. Many applications in computer vision can be formed as image-to-image translation problems such as sketch to face and satellite photos to Google maps. The goal of image-to-image translation is to learn the mapping from a given image in domain $X$ to a specific target image in domain $Y$. Performing such a task requires an understanding of underlying features such that the transformation applies only on the domain-specific part (e.g., the style of a painting) while the domain-invariant part (e.g., the content of a painting) remains unchanged. It is challenging to learn the mapping between two or multiple domains. Recently, many GAN variants have been proposed and provide state-of-the-art solutions to image-to-image translation problems. 

\subsubsection{Pix2pix and Pix2pixHD}
Pix2pix is a supervised image-to-image translation approach proposed by Isola et al. \cite{Isola2017ImagetoImageTW} in 2016. The proposed framework is based on a conditional GAN that takes two images from different domains---one as input, the other as its condition---to perform the translation. It therefore requires paired images to learn the one-to-one mapping. The model consisted of a U-Net-based generator \cite{ronneberger2015u} and a PatchGAN-based discriminator \cite{li2016precomputed}: The U-Net-based generator benefits from the skip connections to pass the vital low-level information shared between the input and output while the PatchGAN-based discriminator breaks the image into patches and focuses on modeling high-frequency structures like edges. The objective function of Pix2pix combines cGAN loss with the L1 norm, which is introduced to enforce correctness at the low frequencies, leading to less burring output images. This model showed excellent results and opened a door to a variety of translation applications such as semantic segmentation, map generation in aerial photography, and colorization of black and white images.

Following the framework, Wang et al. \cite{wang2018pix2pixHD} proposed Pix2pixHD to extend the output resolution from 256$\times$256 to 2048$\times$1024. The authors introduce several crucial changes into the network: a coarse-to-fine generator, a multi-scale discriminators, and a novel adversarial learning objective function that incorporates a feature matching loss to stabilize the training process. 
Extensive evaluation results have shown that the new design advanced both the quality and the resolution of deep image synthesis. However, the framework was still trained in a fully-supervised manner and requires paired training samples.

\figcycleGAN
\subsubsection{CycleGAN}
To overcome the paired image-to-image translation problems, Zhu et al.~\cite{Zhu_2017_ICCV} proposed CycleGAN to address this issue. CycleGAN is trained to learn a mapping between unpaired images from two different domains utilizing two sets of generator and discriminator. As shown in \autoref{fig:cycleGAN}(a), the model consists of two mapping functions $G: X\rightarrow Y$ and $F: Y\rightarrow X$, associating with the discriminators $D_Y$ and $D_X$, respectively. The two sets of generator and discriminator operate symmetrically---the generator $G$ maps the input from domain $X$ to $Y$ while the generator $F$ performed the mapping from $Y$ to $X$. Likewise, the discriminator $D_Y$ distinguishes a translated image $G(x)$ from a real image $y \in Y$ while the discriminator $D_X$ differentiates $F(y)$ from a real image $x \in X$. 

Furthermore, a novel cycle consistency loss was proposed and played a key role in the whole framework. The intuition behind is that the image translation cycle should be able to bring each translated image $G(x)$ back to the original image $x$, i.e., $x\rightarrow G(x)\rightarrow F(G(x))\approx x$, as shown in \autoref{fig:cycleGAN}(b). The authors refer it as \emph{forward-cycle consistency}. Similarly, as illustrated in \autoref{fig:cycleGAN}(c), for each translated image $F(y)$, the two generator $G$ and $F$ should also satisfy \emph{backward-cycle consistency}: $y\rightarrow F(y)\rightarrow G(F(y))\approx y$. 

CycleGAN achieved good results on many translation tasks, such as object transfiguration, collection style transfer and season transfer. Moreover, the proposed cycle consistency loss stimulated several subsequent works in the area of unsupervised image translation. 

\figMUNIT
\subsubsection{UNIT and MUNIT}
UNsupervised Image-to-image Translation (UNIT) \citep{Liu2017UnsupervisedIT} is an unsupervised image-to-image translation framework based on Couple GANs \citep{Liu2016CoupledGA}. It proposes a shared latent space assumption and a weight-share constraint is applied to enforce the shared latent space to generate corresponding images in two domains. However, the performance of UNIT relies on the two domains to have similar patterns and the learned model is unimodal due to the Gaussian latent space assumption. Later, Huang et al.~\cite{Huang2018MultimodalUI} extended UNIT to Multimodal Unsupervised Image-to-image Translation (MUNIT) by revising the shared latent space assumption. Instead of assuming a fully shared latent space as UNIT, the authors postulate that the latent space of images can be decomposed into two: a domain-specific part (i.e., style) and a domain-invariant part (i.e., content). The model consists of two autoencoders as shown in~\autoref{fig:MUNIT}: one encodes the content of the image into a content code and the other encodes its style into a style code. To achieve the generation of multimodal images, MUNIT proposes a training scheme that recombines the encoded content with a randomly sampled style code from the style space of the target domain. 
The trained model therefore produces diverse output based on a given input image by applying different style codes. 
In parallel to MUNIT, Lee et al. \cite{Lee_2018_ECCV} proposed DIRT, which shares the same high-level concept in disentangling the latent space but differs in the way of combining the content and the style code. The following work DIRT++ \cite{lee2019drit} introduced a mode-seeking regularization term to alleviate the mode collapse problem in DIRT, which helped to improve sample diversity.

\figStarGANtwo
\subsubsection{StarGAN}
It is worth mentioning that the methods discussed above are limited to two domains. To tackle this issue, StarGAN~\cite{Choi_2018_CVPR} has been proposed as a unified GAN for multi-domain image-to-image translation using only a single pair of generator and discriminator. Given an input image $x$ and a randomly sampled target domain label $c$, the generator is trained to produce an output image $y$ matching the distribution of the target domain. The authors proposed a simple but effective approach to learn mappings among multiple domains of different datasets by adding a mask vector to the domain label. 
Together with an auxiliary domain classifier on top of the discriminator, StarGAN can therefore perform translation between various domains with a single generator while achieving excellent quality in generated images. 

However, the translation of StarGAN is limited to the local area and the model still learns a deterministic mapping per each domain. Choi et al.  \cite{Choi_2020_CVPR} proposed StarGAN v2 to address the aforementioned problems. To introduce multi-modality to the model, StarGAN v2 replaces the domain labels used in StarGAN with newly proposed domain-specific style codes, which represent diverse styles of a specific domain. The model consists of four modules: a mapping network, a style encoder, a generator, and a discriminator as shown in \autoref{fig:starganv2}. The generator receives an image and a style code as input, where the style code is used to modulate the AdaIN layers in the network. The style code is obtained from either the mapping network or the style encoder. The mapping network learns to transform random Gaussian noise into a style code while the style encoder learns to extract the style code from a given reference image. These two networks are designed to have multiple output branches to provide style codes for a specific domain.
The learned style distribution of each domain is the key for StarGAN v2 to synthesize diverse images over multiple domains. Finally, the multi-task discriminator is trained to 
distinguish whether the input image is a real image or a synthetic one generated by the generator.

Extensive experiments have shown that StarGAN v2 achieved superior results compared to other methods in terms of visual quality, diversity, and scalability.

\section{Application Models for Limited Data}\label{sec4}
In scenarios where data availability is limited, effectively training Generative Adversarial Networks (GANs) becomes a significant challenge. To address this issue, various approaches have been developed to enhance the ability of GANs to learn from limited data while maintaining stability and generalization. This section explores key approaches proposed in the literature to address this challenge. We first examine data augmentation techniques, which artificially expand training data diversity while preserving consistency with the original distribution. We then discuss few-shot learning models, which enable GANs to adapt to new domains with minimal examples by leveraging specialized architectures or transfer learning strategies.
\subsection{Data Augmentation in GANs}
\label{Sec:data-aug-gan}
Even though GANs have shown promising performance in various image synthesis tasks, the framework itself is notorious for requiring large-scale data for stabilized training. Training GANs with limited image data generally results in deteriorated performance and collapsed models due to the overfitted discriminator. As an effective remedy for the data-insufficiency problem, data augmentations have been widely studied and proven to improve the accuracy and robustness of classifiers in limited data regime. However, it is not trivial to apply such technique during GAN training because augmenting training data directly alters the distribution of real images thus mislead the generator. 
Several works have been proposed to address this issue and adapt data augmentations in GAN training.

\figADA

\subsubsection{Training Generative Adversarial Networks with Limited Data}
Karras et al. \cite{karras2020training} proposed Adaptive Discriminator Augmentation (ADA) to mitigate the discriminator overfitting problem while preventing leaking augmentation cues to the generator. The authors argue that even though a previous method from \cite{zhao2021improved} introduced consistency regularization (CR) terms in the discriminator loss to enforce consistency for both real and generated images, it actually opened the door for leaking augmentations to the generator. The effects are thus fundamentally similar to dataset augmentation. In contrast to \cite{zhao2021improved}, the authors remove the CR loss terms and exposed the discriminator \textit{only} to augmented images. Moreover, the augmentations are also applied when training the generator as shown in \autoref{fig:ada}. The design is based on an observation in \cite{bora2018ambientgan}: the generator is able to undo corruptions implicitly and find the correct distribution as long as the corruption process is an invertible transformation of probability distributions over the data space. For example, setting the input image to zero 90\% of the time is inevitable while random rotations chosen uniformly from \{$0^\circ, 90^\circ, 180^\circ, 270^\circ$\} are not. Such augmentations can be referred to as \emph{non-leaking} and allowed decisions on the equality or inequality of the underlying sets by observing only the augmented sets. During training, a pipeline of 18 transformations was applied with a fixed probability value $p \in [0,1]$, indicating the strength of the augmentations. To avoid manual tuning of the augmentation strength, the authors suggest to adjust $p$ dynamically based on the degree of overfitting. The degree of overfitting is quantified by observing the non-saturating loss and turn it into two plausible heuristics
\begin{equation}
r_v = \frac{\mathbb{E}[D_\textrm{train}] - \mathbb{E}[D_\textrm{validation}]}{\mathbb{E}[D_\textrm{train}] - \mathbb{E}[D_\textrm{generated}]}\hspace{5mm} \text{and}\hspace{5mm}%
r_t = \mathbb{E}[\textrm{sign}(D_\textrm{train})]~,
\end{equation}
\noindent where $r = 0$ means no overfitting and $r = 1$ indicates complete ovefitting. Extensive experiments show that with the proposed mechanism the overfitting no longer occured even when training on small datasets.

\figDiffAug

\subsubsection{Differentiable Augmentation for Data-Efficient GAN Training}
In parallel to \cite{karras2020training}, Zhao et al.~\cite{zhao2020differentiable} made similar observations and proposed Differentiable Augmentation (DiffAugment) to tackle the overfitting of the discriminator and train GANs in a data-efficient manner. Despite extensive efforts have been made to find better GAN architectures and loss functions, a fundamental challenge remains: the discriminator tends to memorize the observations as the training progresses. The authors demonstrate that the discriminator suffers from a similar overfitting problem as the binary classifier and provide step by step insights on why dataset augmentation is not effective in GANs. They observe that directly applying the augmentation $T$ to the real data $x$ without other procedures results in learning a different data distribution $T(x)$. This limits the choices of augmentations because any augmentation that significantly alters the distribution of the real images would introduce artifacts to the generated images. To match the generated distribution with the manipulated real distribution, it is intuitive to use the same $T$ on both real and fake samples. If the generator successfully learns the distribution of $x$, the discriminator should fail on distinguishing the real and generated samples as well as their augmentation version. However, this strategy breaks the delicate balance between the generator and the discriminator and leads to an even worse performance. The authors thus conclude that the augmentation has to be applied to both real and fake images for both generator and discriminator training. Moreover, the augmentation $T$ must be differentiable since gradients should be back-propagated through $T$ to the generator. 
Experiments on multiple datasets show that DiffAugment alleviates the overfitting problem and achieves better convergence with simple choices of transformations.

\subsubsection{On Data Augmentation for GAN Training}
Another concurrent work addressing the same issue is Data Augmentation optimized for GAN (DAG) \citep{tran2021data}. Different to previously mentioned methods, 
DAG is based on the Jensen–Shannon (JS) preserving property, which is assured when an invertible transformation is applied.
The framework of DAG consisted of multiple discriminators, where each of them is responsible for a type of transformation. 
The goal of the generator now is to fool all the discriminators simultaneously. It is worthy noting that the generator aims to generate only the original images, not the transformed ones. The generated images are transformed by specified transformations before feeding them to the respective discriminators. As a result, the generator is enforced to produce realistic looking samples with the constraint that their transformed counterparts also look real. The authors provide detailed theoretical analysis to show that the proposed DAG aligns with the original GAN in minimizing the JS divergence between the original distribution and the model distribution. Also, extensive experiments conducted on different GAN models and different datasets show that DAG achieves consistent improvements across these models in the limited data scenario.

\subsubsection{Image Augmentations for GAN Training}
In the research conducted by Zhao et al. \cite{Zhao2020ImageAF}, they reached the same conclusion as earlier studies: it is essential to apply augmentations to both real and generated images during the training of GANs and to both the generator and discriminator. The authors explored the effectiveness of several established augmentation techniques for GAN training and presented a comprehensive analysis of their findings. Moreover, the authors investigate combining augmentation-based regularization techniques with the augmentation strategies and demonstrate that such regularization is not only beneficial but also essential to achieve superior results. Extensive experiments on a broad set of common image transformations 
show that spatial transforms like \emph{zoom out} and \emph{translation} substantially improve the GAN performance when training with balanced Consistency Regularization (bCR). In contrast, \emph{instance noise} cannot improve generation performance. As for regularization techniques, the authors conclude that constrastive loss shows a similar performance to bCR but helps to learn better representations. A new state-of-the-art on Cifar-10 was achieved in this paper by applying both constrastive loss and bCR during training as well as the best augmentation strategy they found.

\subsection{Few-Shot Learning}
\label{sec:few-shot-gan}
Apart from using data augmentations in GANs, one of the other popular trends is few-shot learning. The original goal of few-shot learning is to learn a discriminative classifier where the available data of the target class 
is limited. Recently, a number of work has extended the framework to generative tasks, aiming to generate diverse results while preventing the model from being over-fitted to the few examples or collapsing to a single mode. The approaches to address these issues can be categorized into two main types: 1) Designing novel neural network architectures and training schemes to stabilize the models in low data regimes when training from scratch. 2) Incorporating the transfer learning pipeline to adapt a pretrained GAN on a small target domain.

\subsubsection{Towards Faster and Stabilized GAN Training for High-fidelity Few-shot Image Synthesis}
It is non-trivial to train GANs from scratch in a low data regime that has less than 100 images, even with the help of dynamic data augmentations as discussed in Section~\ref{Sec:data-aug-gan}. The models still suffer from drastic overfitting and mode collapse. Moreover, the computing cost of the state-of-the-art models such as StyleGAN2 \cite{Karras_2020_CVPR} and BigGAN \cite{Brock.25.02.2019} remain to be high, which makes them inapplicable for broader applications. To mitigate these two major pitfalls of GANs, namely, data hunger and high computing cost, Liu et al. \cite{liu2020towards} proposed a light-weight GAN structure for the few-shot image synthesis task. The main contributions are two-fold: First, they redesign the generator structure of StyleGAN and incorporate a novel Skip-Layer channel-wise Excitation (SLE) module to allow faster training. Then, a self-supervised discriminator is introduced to learn more descriptive features and thus, provides more comprehensive signals to stabilize the GAN training. 
The authors reformulate the skip-connection concept from the widely used Residual structure (ResBlock) \citep{he2016deep} with two critical changes: 1) The summation in ResBlock is replaced by channel-wise multiplications between the activations, which reduces the number of parameters for the convolutions by a large margin. 2) The skip-connection between resolutions is applied to a longer range than in the original design, providing stronger gradient signals between layers. These two features also allow the generator to automatically disentangle the content and style attributes like in StyleGAN. As for the self-supervised discriminator, 
 several small decoders are introduced to be optimized together with the discriminator with a reconstruction loss, enforcing the discriminator to extract a more comprehensive representation from the inputs. 
 
 Experiments on multiple datasets demonstrate the effectiveness of the designs and show superior performance compared to the state-of-the-art StyleGAN2 while being efficient with regard to both data and computing cost.

\figFUNIT
\subsubsection{Few-Shot Unsupervised Image-to-Image Translation}
Other than few-shot image synthesis, Liu et al.~\cite{Liu2019FewShotUI} address the few-shot image-to-image translation with a novel network design. The aim of the work is to perform unsupervised image-to-image translation on previously unseen target classes with only a few example images at test time. The author design a training scheme to mimic the few-shot generation capability of humans---the model is exposed to many different object classes during training and is trained to extract appearance patterns from the few examples given in each class. The hypothesis is that the model then learns a generalizable appearance pattern extractor, which can be applied to unseen classes at test time. The proposed model, termed FUNIT, consists of a conditional image generator $G$ and a multi-task discriminator $D$. The generator $G$ takes a content image $\xx$ from object class $c_x$ and a set of $K$ images $\{\mathbf{y}_1,...,\mathbf{y}_K\}$ from object class $c_y$ as input, where $K$ is a small number (e.g., five) and $c_x$ is different from $c_y$. The generator is tasked to extract class-invariant (e.g., object pose) and class-specific (e.g., object appearance) features using two encoders and to produce the output image by modulating the class-invariant latent code with the class-specific one through AdaIN. The output image $\xt$ from $G$ thus should look like an image belonging to object $c_y$ while sharing structural similarity with $\xx$, as illustrated in \autoref{fig:funit}. The multi-task discriminator $D$ is then trained to solve multiple adversarial binary classification tasks simultaneously.

Extensive experiments in various settings with different numbers of $K$ (e.g., $K \in \{1,5,10\}$) shown promising results when translating an input image to an unseen target class. Moreover, the authors also demonstrate that the model performance is positively correlated with the number of object classes available during training. However, due to the training scheme, FUNIT often fails when the appearance of novel objects classes is dramatically  different  from  the training set.

\figsingan
\subsubsection{SinGAN: Learning a generative model from a single natural image}
Contrary to capture the distribution of a set of images, Shaham et al.~\cite{Shaham2019SinGANLA} proposed SinGAN as an unconditional generative model which aims to learn the internal distribution of patches within an image and to produce diverse samples containing the same visual content. In contrast to other single images GAN schemes, SinGAN maintains both the global structure and the fine texture of the training images and thus, is not limited to texture images. The key design of the framework is a pyramid of fully convolutional GANs, which is responsible for capturing the internal statistics of patches at different scales as shown in \autoref{fig:singan}. In detail, the model consists of a pyramid of generators $\{\mathbf{G}_0,...,\mathbf{G}_N\}$, trained against a pyramid of discriminator $\{\mathbf{D}_0,...,\mathbf{D}_N\}$ at different image scales $\{\mathbf{x}_0,...,\mathbf{x}_N\}$, where $\xx_n$, $n=0,\ldots, N$, is a downsampled version of an input image $x$. The generator $G_n$ at each scale is therefore encouraged to generate realistic image samples to the corresponding image $x_n$ regarding the patch distribution. Note that all the generators and discriminators have the same receptive field, which means the effective patch size decreases when going up the pyramid. Moreover, a spatial white Gaussian noise $z_n$ is injected at each scale along with an upsampled version of the image from the coarser scale, adding details that are not generated by the previous scales. 

Various experiments illustrate that SinGAN can be used to solve a variety of image manipulation tasks such as paint-to-image, image editing, or super-resolution from a single image. 

\figoneshotgan
\subsubsection{One-Shot GAN: Learning to Generate Samples from Single Images and Videos}
Similar to SinGAN, Sushko et al.~\cite{Sushko_2021_CVPR} proposed a framework, termed One-shot GAN, which learns to generate samples from one image or one video. However, the authors argue that a patch-based approach such as SinGAN cannot capture high-level semantic properties of the scene and the generated images often suffer from distorting objects and the incoherence between patches. The design of One-shot GAN therefore goes beyond patch-based learning and aims to generate novel plausible compositions of objects in the scene while maintaining the original context of the image. To achieve the goal, two key features are introduced to the model: a novel designed discriminator and a diversity regularization technique for the generator. The new One-shot GAN discriminator consists of two branches, one for judging the content distribution and the other for examining the realism of the scene, as illustrated in \autoref{fig:oneshotgan}. It enforces the generator to produce objects and to combine them in a globally-coherent way. To further regularize the generator, a diversity regularization is introduced to the generator and encourages it to generate perceptually different images. 

Extensive evaluation show that One-shot GAN mitigates the memorization problem in the low data regime and generates images with novel views and object compositions that differ from the training set. Moreover, it improves prior works in both image quality and diversity, and provides the extension to videos.

\subsubsection{A Closer Look at Few-shot Image Generation}
The paper by Zhao et al. \cite{Zhao.2022} analyzes the performance of state-of-the-art generative few-shot learning methods based on the fine-tuning of networks.
The authors first analyze the ability of the architectures to generate quality images by proposing a systematic test for verifying the quality and diversity of generated data. The quality of the network is assessed through the use of a binary classifier that receives two sets of images (from the source and target domains) and returns a probability $\nicefrac{p_t}{1-p_t}$ that the input belonged to the target domain or source domain, respectively. The diversity is 
measured by an intra-cluster LPIPS (intra-LPIPS, for LPIPS see Section~\ref{sec:metrics}) \cite{Ojha.2021} that evaluates the ``perceptual distance between two images". Some networks like TGAN \citep{Wang.2018}, ADA \citep{Karras.2020}, BSA \citep{Noguchi.2019}, and FreezeD \citep{Mo.25.02.2020} are evaluated. 
Although the methods evaluated can achieve acceptable quality in the target domain, the results show that, on the one hand, some architectures tend to preserve the diversity of the source context at the expense of the quality of the generated images of the target context. On the other hand, other architectures achieve similar quality in the target domain but with a dramatically lower diversity rate.
In this regard, a method for decreasing the degree of diversity degradation based on dual contrastive learning is presented in the second part of the work. The basic idea is to maximize the mutual information between the source and target image features originating from the same input noise $z_i$ and pushing away the generated images on the source and target domain that use different noise input.
For this purpose, the agreement between positive pairs is maximized, i.e., pairs of images generated in the source and target domains with the same input noise. The loss function thus includes two terms in addition to the opposing loss function: one from the generator's view and the latter from the discriminator's view.

\subsubsection{Few-Shot Generative Model Adaption via Relaxed Spatial Structural Alignment}
Xiao, Li et al. addressed the problem of few-shot learning by proposing a so-called \emph{relaxed spatial structural alignment} (RSSA) \cite{Xiao.2022} method to calibrate the generative model during the training/adaptation phase in the target domain.
The strategies by the authors focus 1) on preserving the prior structures of images from the source domain and transferring them to the target domain, and  2) to speed up the training process by compressing the latent vector and facilitating cross-domain alignment. 
The first aspect is achieved by means of a \emph{cross-domain spatial structural consistency loss},  consisting of the \emph{self-correlation consistency loss} $L_\mathrm{scc}$ that constrains the inherent structure of the images and the \emph{disturbance correlation consistency loss} $L_\mathrm{dcc}$ that constrains its variation within a disturbance limit.
These losses help the alignment of structural information between the synthesis image pairs of the source and target domains.
The second strategy presented is focused on compressing the original latent space into a subspace closer to the target domain. The latent vector of the $l$-th layer $w_j^l$, obtained from the input noise $z_j$, is modulated and projected via the least-square method into a $\mathcal{X}^l$ subspace.  
The subspace $\mathcal{X}^l$ represents one of the $n$ samples of the target domain that are transformed from target domain $\{x_i\}_{i=1}^{n}$ to the source space $W^+$ of $G_s$, where $G_s$ represents the source domain generator.
The experiments show that the RSSA method effectively improves the adaptation of generative models with limited data by maintaining the spatial structure of the original images and accelerating convergence through latent space compression. However, the method struggles with highly abstract domains (e.g. Modigliani-style portraits) where extreme distortions in facial proportions make structural alignment less effective.

\subsubsection{Few-Shot adaptation of GANs}
In 2020, Robb et al. presented FS-GAN \cite{Robb.2020}, a method for adapting GANs to few-shot learning scenarios.
The basic idea is to restrict the space of trainable parameters to a small number of highly representative features and modulate these orthogonal features.
The method first uses a singular value decomposition (SVD) to the weights of a pre-trained GAN. The SVD is applied separately at every layer of the generator and discriminator.
Then, the domain adaptation is performed by freezing pretrained left/right singular vectors and optimizing the singular values using the standard GAN objective function. RSSA generates realistic and visually consistent images, effectively preserving the spatial structures of the source domain while capturing the characteristics of the target domain. Tests with different datasets under different low-data regimes show that the method achieves the highest IS metrics, ensuring diverse and high-quality generated images.

\subsection{GAN Inversion}
\label{sec:gan-inversion}
As an emerging technique to interpret a GAN’s latent space, \emph{GAN inversion} servers as a proxy between real and fake image domains and plays an essential role in enabling powerful pretrained GAN models like StyleGAN and BigGAN for various downstream applications. 
Several works have been proposed to exploit the learned latent space of GANs, identity new interpretable control directions, and offer insights on the limitations in image generation. By leveraging the rich information encoded in a pretrained GAN, GAN inversion not only provides a flexible framework for tasks like image editing but also largely reduced the need for data and computing power compared to training GANs from scratch.

\subsubsection{GANSpace: Discovering Interpretable GAN Controls}
Härkönen et al. presented GANSpace \cite{Harkonen.2020}, a technique that enables control of the image synthesis process using principal component analysis (PCA).
The work is based on the idea that principal components of the features tensor on the early layers of a GAN can represent factors of variation. Therefore a layer-wise perturbation along the principal direction can produce more interpretable control in the synthesis process and more variety in the generated data. The method is applied to two architectures: StyleGAN and BigGAN.

For the StyleGAN architecture \cite{Karras_2019_CVPR}, PCA is applied to $N$ intermediate latent space representation $\mathcal{W}$, selecting $N$ random vector $z_{1:N} \in \mathcal{Z}$. This PCA operation gives a basis $V$ for $\mathcal{W}$.
Furthermore, using the basis $V$ for $\mathcal{W}$, a new image, indicated as intermediate latent representation $w \in \mathcal{W}$, can be edited with varying the PCA coordinates $h$ before the feeding to the synthesis network.
On the other hand, regarding the application to BigGAN \cite{Brock.25.02.2019}, since it is not possible to work directly with the latent vector distribution $z$, the authors performed PCA at an intermediate layer $i$ of the network. 
Also in this case, $N$ random latent vectors $z_{1:N}$ were sampled and then fed to the network. The $N$ intermediate feature tensors $y_{1:N}$ at layer $i \in \{1...N\}$ 
are then used to calculate PCA.
 Finally, the basis is transferred to latent space using linear regression.
Given a new image, editing is made possible for both methods presented by changing the PCA $h$ coordinate before passing it to the synthesis network.

\subsubsection{Seeing What a GAN Cannot Generate}
To visualize and understand the semantics concepts that a GAN generator cannot generate, Bau et al. \cite{Bau.2019} investigate the mode collapse at both the distribution and instance level and present a method for inverting a GAN focusing on the inversion of the single layers instead the entire generator. 
First, the study calculates the deviation between true and synthetic distributions. 
It consists of segmenting the generated and target data to identify which objects are omitted from the generator. All the training and generated image are segmented, and the total area in pixels for each object class, together with means and covariance statistics, are measured.
The image segmentation statistics are then summarized by introducing a \emph{Fr\'echet segmentation distance} (FSD), a modification of the Fr\'echet inception distance (cf. Sec.~\ref{sec:metrics}).
Second, the authors focus on the instance level, looking at how particular object classes are omitted by the generator. 
In this phase, a layer-wise network inversion is performed. The generator $G$ is decomposed into layers $G = G_f(g_n(\cdots((g_1(z))))$, where $g_1,..., g_n$ are several early layers of the generator and $G_f$ groups all the later layers of the $G$ together. A neural network $E$, which approximately inverts the generator $G$, is then developed to estimate an initial latent vector $z_0=E(x)$.  
The initial latent vector $z_0$ and its intermediate representation $r_0 = g_n(...(g_1(z_0)))$ are then used to perform a layer-wise optimization to find an intermediate representation $r^\ast$ able to generate image $G_f(r\ast)$ closely similar to the target image $\mathrm{x}$. Experimental results show that GAN generators tend to omit specific object classes entirely rather than rendering them with low quality or distortion. For instance, in the case of scene generators trained on datasets such as LSUN Bedrooms and LSUN Churches, objects like people, fences, and architectural details are systematically absent from the generated images. The FSD highlights that architectures like StyleGAN better match target distributions than older models like WGAN-GP but still exhibit omissions. At the instance level, the layer-wise inversion method effectively identifies these omissions by reconstructing real images and exposing the semantic gaps in the generator’s latent space.

\subsubsection{In-Domain GAN Inversion for Real Image Editing}
Zhu et al. \cite{Zhu.2020} addressed the topic of GAN inversion by proposing an approach `in-domain', where the inversion process does not focus only on reconstructing the target image using the pixel values but it also ensures that the latent code includes semantic knowledge. For this purpose, a domain-guided encoder with a domain-regularized optimization is introduced. The domain-guided encoder is illustrated in \autoref{fig:inDomainGAN} (a). The GAN's generator and discriminator are involved in training the encoder $E$ to spread semantic information. During the training, it receives real images as input instead of synthetic ones and returns a latent vector $z^{enc}$ fed into the GAN generator. The discriminator then evaluates the generated images to ensure they were realistic enough.
Domain-regularized optimization ensures a better correspondence at the pixel level between the target image and the reconstructed image. The proposed approach delivers high-quality image reconstructions and facilitates advanced image editing tasks. The reconstructed images maintain pixel-level fidelity and semantic alignment with the target, ensuring meaningful and coherent outputs. Precision-recall evaluations confirm that the latent codes retain robust semantic information, enabling tasks such as attribute-based manipulations of images. For example, edits to attributes like pose, expression, and the addition or removal of eyeglasses were achieved with minimal distortion to other image details. Furthermore, the interpolation of two images using the method produced smooth transitions that were visually plausible and semantically consistent.

\figInDomainGAN

\subsubsection{Image2StyleGAN and Image2StyleGAN++}
Image2StyleGAN \citep{Abdal.2019} and Image2StyleGAN++ \citep{Abdal.2020} are two works by Abdal et al. based on the study of the latent space of the StyleGAN architecture.
In Image2StyleGAN, the authors show how to embed a given image into a latent space of StyleGAN and how it is possible, by performing basic operations on vector in the latent space, to perform image editing operations like image morphing, style transfer, and expression transfer. For this purpose, an extended latent space $W^+$ consisting of a concatenation of 18 different 512-dimensional vectors $\mathrm{w}$ is considered.

In the following work \cite{Abdal.2020} the authors improve the quality of the images generated, allowing local control over the embedding process.
The main improvements are mainly an extended embedding algorithm into the $W^+$ space allowing local modifications and a new optimization strategy to restore high-frequency features.
The extended embedding algorithm is a gradient-based optimization algorithm to iteratively update the synthesized image, initialized by means of a latent code from two latent spaces. The algorithm's inputs are a couple of images $x$ and $y$ and some spatial masks ($M_s$,$M_m$, $M_p$).
The optimization strategy developed aims to improve the quality of synthetic images using the space $W^{+}$, encodes as much meaningful information as possible and the Noise space $N_{s}$ encoding high frequency details.
The authors found that an alternating optimization strategy between the vectors $w \in W^{+}$ and $n \in N_s$ (optimizing $w$ while $n$ is fixed and then optimizing $n$ while keeping $w$ fixed) provides a better performance than a joint optimization of the vectors.

\subsubsection{Encoding in Style: A StyleGAN Encoder for Image-to-Image Translation}
Style control by handling the $W^+$ vector and using the StyleGAN generator is also used in the work of Richardson et al. \cite{richardson2021encoding}. With Pixel2style2pixel (pSp), they present an image-to-image translation framework where an encoder can perform GAN inversion without the need for optimization.
As shown in \autoref{fig:pSp}, the target image is first fed to the encoder. The encoder $\mathrm{E}$ is composed of a feature pyramid with three levels. Each level represents a different level of detail (coarse, medium, and fine), roughly corresponding to three levels of the StyleGAN style inputs. Then, a small mapping network called map2style is trained for each level to extract the learned styles from the corresponding features map. Finally, the styles are fed into the StyleGAN generator to generate the output image. 

The loss function used for the encoder training is a weighted combination of several objectives. On top of the pixel-wise L2, an LPIPS loss is used to learn the perceptual similarities. A regularization loss encourages the encoder to output latent style vectors close to the average latent vector. Finally, a loss based on similarity is used for preserving the input identity.
The combination of the proposed encoder with the StyleGAN decoder makes it possible to create a generic framework for image-to-image translation tasks.
The results show improvements in applications such as StyleGAN inversion, facial frontalization, and conditional image synthesis. In particular, the encoder achieves high-fidelity reconstructions with enhanced identity preservation, as demonstrated by comparisons with state-of-the-art approaches.

\figpSp

\figdatasetgan

\subsubsection{DatasetGAN}
Recent research has shown that GANs encode rich semantic information within their latent space, even in an unsupervised setting. With this foundational observation, Zhang et al.~\cite{Zhang2021DatasetGANEL} proposed DatasetGAN, a framework that requires only a few labeled examples to produce an infinite number of high-quality, semantically segmented images. DatasetGAN builds upon StyleGAN with an additional \emph{Style Interpreter} to decode the intermediate latent feature maps into target semantic labels. While StyleGAN is considered as a rendering engine in the framework, the Style Interpreter acts as a label-generating branch, allowing DatasetGAN to synthesize image-annotation pairs. The authors propose to upsample all feature maps to the highest output resolution and to concatenate them together to serve as the input to the Style Interpreter, which was a three-layer MLP classifier acting on top of each feature vector to predict target labels as shown in \autoref{fig:datasetgan}. Due to the high dimensionality (5056 dimensions) and high spatial resolution (1024 dimensions) of the concatenated feature map, random sampling is performed during the training and the final Style Interpreter is an ensemble of $N$ classifiers.

The proposed Style Interpreter needs only a few annotated examples for achieving a good accuracy, therefore it is possible to label images in extreme detail and generate large-scale datasets with rich segmentations, requiring minimal human effort. The authors showcase that together with a simple filtering mechanisms, DatasetGAN outperforms all semi-supervized baselines in seven image segmentation tasks and is comparable to fully supervized methods with only a handful of annotated data.

\subsubsection{BigDatasetGAN}
Despite the success of DatasetGAN, it is non-trivial to adapt it to conditional generative models. To this end, Li et al. \cite{li2022bigdatasetgan} proposed BigDatasetGAN to extend DatasetGAN to work on BigGAN \cite{Brock.25.02.2019} and VQGAN \cite{Esser2021TamingTF}, which are two conditional generative models pretrained on ImageNet. The two chosen networks have largely different architectures and training approaches: BigGAN is fully convolutional and trained with standard adversarial losses. On the other hand, VQGAN utilizes an autoregressive transformer to model the composition of context-rich visual parts in latent space along with convolutional encoder and decoder networks.
The aim is to learn a \emph{feature interpreter} $\mathcal{S}$ performing segmentation based on given classes.
The authors propose to group features of different spatial resolutions into three levels---high, mid, and low.
The feature maps at different levels are then upsampled and concatenated in a progressive fashion, which greatly reduces the memory cost and preserves more contextual information compared to DatasetGAN. For VQGAN, the features of the transformer and the decoder are also included in the feature set for producing segmentation maps. One notable difference between BigGAN and VQGAN is that the network design of VQGAN allows it to embed images other than its own generated samples with excellent reconstruction fidelity, while there are yet no satisfactory encoders for BigGAN. Therefore, the annotated BigGAN samples are used to train both BigGAN and VQGAN. Extensive experiments demonstrate that the synthesized datasets generated by BigDatasetGAN improved over standard ImageNet pre-training on several datasets across various downstream tasks such as detection and segmentation.

\figgpunit

\subsubsection{Unsupervised Image-to-Image Translation with Generative Prior}
Although unsupervised image-to-image translation has been studied extensively in recent years, big challenges remain in transforming between complex domains with drastic visual discrepancies. To mitigate the common failure in previous works in this regard, Yang et al.~\cite{yang2022unsupervised} proposed to leverage the generative prior from pretrained class-conditional GANs and termed their framework Generative Prior-guided UNsupervised Image-to-image Translation (GP-UNIT). The key insight is that pretrained class-conditional GANs like BigGAN \cite{Brock.25.02.2019} generate images with a high degree of content correspondence (e.g., having the same pose) when given the same latent code. The authors therefore propose to mine the unique prior embedded in the class-conditional GAN and use them as guidance in downstream translation tasks. The framework consists of two stages: 1) generative prior distillation and 2) adversarial image translation as shown in \autoref{fig:gpunit}. The goal of the first stage is to learn robust cross-domain correspondences at a high semantic level---a content encoder $E_c$ is trained to extract shared coarse-level features among generated images of different classes but conditioned on the same latent code. In the meantime, a decoder $F$ aims to reconstruct the input image $x$ based on its content feature $E_c(x)$ and a style feature encoder $E_s(x)$, ensuring the disentanglement of the desired content feature.
The trained $E_c$ is then deployed in the second stage to measure the content similarity. The second stage follows a standard style transfer paradigm together with a novel dynamic skip connection module to build finer adaptable correspondences at multiple semantic levels. The proposed dynamic skip connection module passes the middle layer of $E_c$ directly to the generator, while predicting masks $m$ to select the valid elements for building the fine-level content correspondences that cannot be characterized solely by the abstract content feature. 

The authors showcase that  GP-UNIT surpasses the state-of-the-art image-to-image translation methods on several datasets regarding image quality and diversity, even for challenging and distant domains.

\subsubsection{GAN Dissection: Visualizing and Understanding Generative Adversarial Networks}
The work of Bau et al.~\cite{Bau.2018} presents a framework for visualizing and understanding the internal representations of a GAN generator.
The method investigates how objects (like trees or tables) are internally encoded in the GAN generator and which variables cause the generation of these objects.
Image~\ref{fig:GAN-Dissection} shows the two phases of the proposed framework.
In the first phase, called \emph{Dissection} (\autoref{fig:GAN-Dissection}(a)), the authors want to know if a specific unit $\mathbf{r}_{u,\text{P}}$ encodes a semantic class such as a tree. For this purpose, the units are selected by looking at the correlation level between the feature map generated by the single unit $u$ and the segmented region representing the object $c$ in the generated image $x$.
Once the units that are responsible for generating the object are identified, the second phase, called \emph{Intervention} (\autoref{fig:GAN-Dissection}) asks which of these are responsible for triggering the rendering of them. For this reason, the units of $U$ are forced to switch on and off. 
The causality is then measured by comparing the object's presence in the two synthesized images (with ablated units and forced-inserted units) and averaging the effect over all locations and images.

\figGANDissection


\section{Metrics}
\label{sec:metrics}
Numerous metrics have been presented for evaluating GAN performance. In this section, some of the commonly used metrics are briefly introduced.
\begin{itemize}
 \item \textbf{FID}: The \emph{Fr\'echet inception distance} (FID) is one of the most widely used metrics for evaluating GANs. The metric uses the features generated by the Inception network \cite{szegedy2016rethinking} with real and generated data to calculate the Fr\'echet distance between the two distributions, modeled as a multidimensional Gaussian distribution with mean $\mu_r, \mu_g$ and covariance $\mathbf{C}_r$, $\mathbf{C}_g$. A lower FID indicates a smaller distance between the generated and real data distribution.

 \item \textbf{LPIPS}: The \emph{Learned Perceptual Image Patch Similarity} (LPIPS) distance measures perceptual similarity using deep network activations. The normalized embeddings are used to measure the similarity between two images due to the calculation of L2 distance. The networks commonly used for the metric are SqueezeNet \cite{iandola2016squeezenet}, AlexNet \cite{krizhevsky2012imagenet}, and VGG \cite{simonyan2014very}. The lower the value of LPIPS, the more perceptually similar are the two analyzed images.
 
 \item \textbf{IS}: The \emph{Inception Score} (IS) is a metric that measures the visual quality and diversity of the generated images using the Inception-V3 network \cite{szegedy2016rethinking}. The generated images are fed to the Imagenet pre-trained version of the network. The output is used to calculate the KL-divergence between the conditional class distribution and the marginal class distribution.

\item  \textbf{Precision \& Recall}: In discriminative models, precision measures the fraction of relevant retrieved instances among the retrieved instances, while recall measures the fraction of retrieved instances among the relevant instances. In the context of generative models, the two metrics were introduced in \cite{Lucic.2018}. 
The authors present a toy dataset, a manifold of convex polygons, where the distance from samples to the manifold is used to calculate precision and recall. The precision is high if the samples from the generative model are close to the manifold. Similarly, the recall is high when the model can generate data instances close to any manifold samples.
\end{itemize}

\section{Experiment and Result Analysis}\label{sec6}
We selected some of the GAN architectures proposed in the previous sections and tested them under different stress conditions. Every network was trained with different datasets and different levels of data scarcity.
\subsection{Datasets}
Different subsets were created using five public datasets.
The subsets were limited by the number of instances per class but also by the total number of classes available. An overview of the subsets created is shown in \autoref{tab:sum_datasets}.
\begin{itemize}
\item \textbf{Imagenet:}
The Imagenet dataset contains more than 14 million images annotated according to the Wordnet hierarchy. Two subsets, (A) and (B), containing 1,000 classes and 1\% and 10\%, respectively, of the original number of instances per class were used. The two subsets were downloaded from the official SimCLR repository\footnote{\url{https://github.com/google-research/simclr/tree/master/imagenet\_subsets}}.
\item \textbf{AFHQ:}
AFHQ is a dataset of animal faces representing 15,000 high-quality images divided into three classes: cat, dog, and wildlife. In the two subsets, the number of instances per class was limited to 20 in (C) and 200 in (D). The number of classes remained unchanged.
\item \textbf{MIT Scene Parsing:}
MIT Scene Parsing is a dataset for training and evaluating scene parsing algorithms. The dataset, a subset of the ADE20K dataset, contains approximately 150 semantic categories such as sky, road, grass, etc. Three subsets were created by limiting the number of classes and the number of instances per class. In particular, the classes were limited to the following six: bed, building, cabinet, car, chair, and tree. The number of instances per class was limited to 20 in (E), to 200 in (F), and kept unchanged in (G).
\item \textbf{CelebAMask-HQ:}
CelebAMask-HQ is a dataset containing approximately 30,000 face images. Extending the CelebA-HQ dataset, this one differs from the first by the presence of semantic class maps. Also in this case, the subsets were realized by limiting the number of classes to seven and reducing the number of instances per class to 17 in (H) and 161 in (I).
The classes analyzed were: Eyebrows, Eyeglasses, Hair, Hat, Mouth, Nose, and Skin.

\item \textbf{Animal Faces:} 
The dataset is composed of the carnivorous animal classes from ImageNet, built by Liu et al. \cite{Liu2019FewShotUI}. It contains in total 117,574 animal faces distributed across 149 classes, where the classes are further split into a source class set (119) and a target class set (30) for the image-to-image translation task. Two subsets (K) and (L) were created based on the full dataset, where 10 classes from the source set and 5 classes from the target set were randomly selected. Moreover, in setting (K), the number of available images in each class was further reduced to 20, mimicking an possible extreme case in a real-world scenario.
\end{itemize}

\begin{table}[t]
\begin{center}
\begin{minipage}[t]{1.0\linewidth}
\centering
\caption{Summary of the dataset settings.}
  \label{tab:sum_datasets}

\footnotesize{

\begin{tabular}{ |c|c|c|c|c|c|c|} 
\hline
Dataset &Image &Mask  & Classes (Train/Val)  &Mode  &Train &Val  \\
\hline
\multirow{2}{8em}{ImageNet} &   &    & \multirow{2}{6em}{1,000 / 1,000  } & (A)  &12,811 &\multirow{2}{3em}{128,116}  \\ 
                            & V &   &  & (B)  &128,110 & \\ 
\hline
\multirow{2}{8em}{AFHQ}     &   &     & \multirow{2}{3em}{ 3  / 3  } & (C)   &60 &\multirow{2}{2em}{1463} \\ 
                            & V &   &  & (D)    &600 &\\ 
\hline
\multirow{3}{8em}{MIT SceneParsing}     &   &    & \multirow{3}{3em}{ 6  / 6  }  & (E)   &120 & \multirow{3}{2em}{1,461} \\ 
                            & V & V   &   & (F)  &1,200 & \\ 
                            &  &    &  & (G)   &14,735 & \\ 
\hline
\multirow{3}{8em}{CelebAMask-HQ}     &   &    & \multirow{3}{3em}{ 7  / 7  } & (H) &119 & \multirow{3}{2em}{3,162} \\ 
                            & V & V   &  & (I)   &1,127 & \\ 
                             &  &    &  & (J)   &19,451 & \\ 
\hline
\multirow{3}{8em}{Animal Faces}     &   &    & 10  / 5  & (K)  &200 &100 \\ 
                                    & V &    & 10  / 5   & (L)  &8,018 &4,019 \\
                                    &  &    & 119  / 30   & (M)  &93,404 &24,080 \\
\hline

\end{tabular}
}

\end{minipage}
\end{center}
\end{table}

\subsection{Experimental Design}
Six network architectures from three important image generation tasks---image synthesis, semantic image synthesis, and image-to-image translation---were selected for evaluation. To analyze how the state-of-the-art models powered by large-scale datasets cope with data scarcity, we chose the most commonly used architectures---BigGAN \cite{Brock.25.02.2019} and StyleGAN2 \cite{Karras_2020_CVPR} to represent conditional and unconditional image synthesis; SPADE \cite{park2019semantic} and SEAN \cite{zhu2020sean} for semantic image synthesis; and StarGAN v2 \cite{Choi_2020_CVPR} and FUNIT \cite{Liu2019FewShotUI} for image-to-image translation. Note that most of the networks mentioned in Section~\ref{sec4} were not selected for evaluation because, albeit relaxing the need for data, these networks did not perform on par with the state-of-the-art models.

Each of the six architectures was trained using various datasets and configurations. For the image synthesis task, BigGAN was trained on subsets of Imagenet and CelebMask-HQ, while StyleGAN utilized the CelebMask-HQ and MIT Scene Parsing subsets. Furthermore, we evaluated the performance of BigGAN and StyleGAN with two famous data augmentation techniques for image synthesis, ADA and DiffAugment, as mentioned in Section~\ref{Sec:data-aug-gan}. This was done to assess how effectively these methods could reduce reliance on large datasets.
As for the semantic image synthesis task, datasets providing semantic information were used. For this reason, both SPADE and SEAN were trained using the subsets of MIT Scene Parsing and CelebAMask-HQ. Finally, for the image-to-image translation task, the StarGAN v2 architecture was trained with AFHQ, CelebAMask-HQ, and Animal Faces subsets, while FUNIT \cite{Liu2019FewShotUI} was trained with a modified subset of Animal Faces.

\subsection{Evaluation}
The performance of the trained networks was evaluated using FID metrics. Specifically, CleanFID \cite{parmar2022aliased} was used to measure the difference between real and synthetic data distribution.
The distribution of synthetic data varies depending on the task addressed. In the case of the image synthesis task, 50,000 synthetic instances were considered. In the semantic image synthesis task, on the other hand, the synthetic images were generated from the semantic maps contained in the validation set. Finally, for the image-to-image translation task, 25,000 synthetic images were considered.

\subsubsection{Image Synthesis}
We present the quantitative evaluation results under different settings in \autoref{tab:fid_big} and \autoref{tab:fid_sty}. The qualitative results of BigGAN \cite{Brock.25.02.2019} is shown in \autoref{fig:epbiggan}, where the quality of the sampled images were mostly poor and do not contain objects of the target classes. Despite the poor performance in the limited data regime, it is observed that images from both settings (A) and (B) show signs of the common attributes of the target class. For example, the last three rows in both settings are mostly blue images, reflecting the color of the sea.
Also, it is evident that BigGAN performed better with a larger number of training images, as greater variations were observed in the images generated under setting (B) compared to setting (A). This trend is further supported by the outcomes when DiffAugment was utilized. However, it is important to note that while both the quantitative and qualitative results of BigGAN with DiffAugment demonstrate considerable improvement over those without augmentation, they still fall short of expectations, emphasizing the limitations of these augmentation techniques.

\begin{table}[t]
\begin{center}
\begin{minipage}[t]{1.0\linewidth}
\centering
\caption{Results of the image synthesis method, BigGAN, in FID.}
\label{tab:fid_big}
\footnotesize{
\begin{tabular}{c|c|c|c|c|} 
\hhline{~|-|-|-|-|}
&\multicolumn{4}{|c|}{BigGAN} \\ 
\hhline{~|-|-|-|-|}
&\multicolumn{2}{|c|}{ImageNet} &\multicolumn{2}{|c|}{CelebAMask-HQ} \\ 
\hhline{~|-|-|-|-|}
&(A) & (B)  &(H) &(I)  \\ 
\hhline{~|-|-|-|-|}
&217.90 &154.61  &\textbf{278.97} &220.28   \\ 
\hhline{~|-|-|-|-|}
 +DiffAugment &\textbf{195.49} &\textbf{129.56}  &390.61 &\textbf{130.74}   \\ 
\hhline{~|-|-|-|-|}
\end{tabular}
}
\end{minipage}
\begin{minipage}[t]{1.0\linewidth}
\centering
\caption{Results of the image synthesis method, StyleGAN2, in FID.}
\label{tab:fid_sty}
\footnotesize{
\begin{tabular}{c|c|c|c|c|c|c|} 
\hhline{~|-|-|-|-|-|-|}
 & \multicolumn{6}{|c|}{StyleGAN2}       \\ 
\hhline{~|-|-|-|-|-|-|}
 &\multicolumn{3}{|c|}{MIT Scene Parsing} &\multicolumn{3}{|c|}{CelebAMask-HQ} \\ 
\hhline{~|-|-|-|-|-|-|}
& (E) &(F) &(G) &(H) &(I) &(J) \\ 
\hhline{~|-|-|-|-|-|-|}
 & 274.14 &90.56 & 21.08 &207.88 &183.98 &18.43 \\ 
 \hhline{~|-|-|-|-|-|-|}
 +ADA &   \textbf{233.99} &\textbf{59.79} & \textbf{16.09} & 37.71 &16.67 & 12.24 \\ 
 \hhline{~|-|-|-|-|-|-|}
 +DiffAugment &   250.88 &62.86 & 16.22 & 53.83 & 42.83 & 11.44 \\ 
\hhline{~|-|-|-|-|-|-|}
\end{tabular}
}
\end{minipage}
\end{center}
\end{table}

Moreover, examining cases (H) and (I), it can be seen in \autoref{fig:epbiggan_celeb} that the models trained with the CelebAMask-HQ dataset provide better quality results than Imagenet. This can be partly attributed to the dataset's inherent characteristics that represent only human faces, unlike the different classes represented in Imagenet. But again, in both cases, a mode-collapse of the network can be observed.
In case (H) an initial composition of faces can be recognized although in the presence of numerous artifacts. However, the case (I) shows a collapse of the network and compositional structure of the image although in the presence of fewer artifacts.
Also, it is noteworthy that the use of DiffAugment resulted in a significantly poorer performance in case (H) in comparison to the version without augmentation. This highlights that data augmentation techniques may not always be advantageous, especially when the dataset is limited (e.g., having only 100 samples).

Similar to BigGAN, we observe a significant degradation in FID scores for StyleGAN2~\cite{Karras_2020_CVPR} across both datasets as the number of training samples decreases. This trend is also evident in the qualitative evaluation. As shown in \autoref{fig:epsty}, images generated under setting (G) display a reasonable object-scene composition, although with slight distortions in structural details. In contrast, setting (F) shows the model failing to generate recognizable objects, while setting (E) degrades further, producing only colorful patches. In this regime, techniques like ADA and DiffAugment fail to yield noticeable improvements. However, in setting (F), both methods offer some enhancement in scene composition despite the presence of substantial distortions.

Compared to the MIT Scene Parsing dataset, models trained on the CelebAMask-HQ dataset seem to preserve spatial relationships better under similar training conditions. We hypothesize this is due to the lower compositional complexity of facial structures compared to natural scenes. As shown in \autoref{fig:epsty}, setting (J) generates high-quality samples visually comparable to real images. In contrast, settings (H) and (I) fail to produce coherent facial structures. Notably, although setting (I) suffers from evident mode collapse, it appears slightly better conditioned than setting (H).
Mode collapse is also visible in models trained with ADA and DiffAugment under setting (H), whereas setting (I) retains some semantic structure of faces despite noticeable distortions.
The results from both BigGAN and StyleGAN2 reaffirm that state-of-the-art generative models relying on random noise inputs perform best when trained on large-scale datasets. This is particularly evident in their FID scores. While data augmentation techniques like ADA and DiffAugment consistently provide quantitative improvements, their effectiveness remains limited in low-data regimes, where generated images often still suffer from visual artifacts and distortions.

\figepbiggan

\figepbigganCeleb
\figepsty

\subsubsection{Semantic Image Synthesis}

Unlike image synthesis benchmarks, semantic image synthesis methods utilize additional, conditional information provided by semantic masks, which relax the need for data by a large margin. As shown in \autoref{fig:epspade}, we can clearly observe the outline of generated objects in both datasets despite the extremely limited available samples in settings (E) and (H). 
When training with sightly more images such as settings (F) and (I), SPADE \cite{park2019semantic} delivered more visually plausible results than StyleGAN2 \cite{Karras_2020_CVPR}, which is also reflected in their FID scores. We believe that the faster converge of SPADE is due to incorporating the additional semantic information, which provides the cue for layout and spares the network capacity from modeling the global spatial relationship. 

\begin{table}[t]
\begin{center}
\begin{minipage}[t]{1.0\linewidth}
\centering
\caption{Results of the semantic image synthesis method, SPADE, in FID.}
\label{tab:fid_semantic_SPADE}
\footnotesize{
\begin{tabular}{ |c|c|c|c|c|c|c|c|c|} 
\hline
\multicolumn{6}{|c|}{SPADE}        \\ 
\hline
\multicolumn{3}{|c|}{MIT Scene Parsing} &\multicolumn{3}{|c|}{CelebAMask-HQ}  \\ 
\hline
 (E) &(F) &(G) &(H) &(I) &(J)  \\ 
\hline
181.08 &80.82 &58.01 &79.64 &57.23 &44.02  \\ 
\hline
\end{tabular}
}
\end{minipage}
\begin{minipage}[t]{1.0\linewidth}
\centering
\caption{Results of the semantic image synthesis methods, SEAN, in FID.}
\label{tab:fid_semantic_SEAN}
\footnotesize{
\begin{tabular}{ |c|c|c|c|c|c|c|c|c|} 
\hline
\multicolumn{6}{|c|}{SEAN}       \\ 
\hline
\multicolumn{3}{|c|}{MIT Scene Parsing} &\multicolumn{3}{|c|}{CelebAMask-HQ} \\ 
\hline
(E) &(F) &(G) &(H) &(I) &(J) \\ 
\hline
163.10 &79.78 &45.33 &101.48 &45.56 &21.85 \\ 
\hline
\end{tabular}
}
\end{minipage}
\end{center}
\end{table}

\autoref{fig:epssean} displays the outcomes achieved using SEAN, which aligns with the previous observations made regarding SPADE.
In this case, the performances obtained in quantitative terms shown in \autoref{tab:fid_semantic_SPADE} and \autoref{tab:fid_semantic_SEAN} are better in mostly all the cases analyzed. From the qualitative analysis of the synthesized images, a higher level of detail can be observed, especially in cases (I) and (J), presumably due to the improved normalization technique that better controls individual semantic regions.
However, we also observe that when the whole dataset is accessible by the model, like in settings (G) and (J), StyleGAN v2 achieves a better qualitative and quantitative performance than SPADE and SEAN. We hypothesize that StyleGAN2 has more parameters and a higher degree of freedom and, therefore, benefits more from a larger training set.

\figepspade
\figepsean


\subsubsection{Image-to-Image Translation}
The quantitative results obtained for the image-to-image translation task are presented in \autoref{tab:fid_i2i1} and ~\autoref{tab:fid_i2i_FUNIT}.
We evaluated the images synthesised by StarGAN v2 obtained in both latent-guided and reference-guided modes.
The images related to the dataset AFHQ and animal Faces are shown in \autoref{fig:epsstargan_afhq} and \autoref{fig:epsstargan_animals}, respectively. For both cases, similar performances to BigGAN are observed. Also in these cases the image quality is deficient, and only some attributes of the target class can be recognized.
For the CelebAMask-HQ dataset shown in \autoref{fig:epsstargan_celeb}, better performances are observed for cases (H) and (I) due to, in our opinion, the smaller domain shift between the classes in the dataset.
However, a massive mode collapse of the network is observed in all cases representing a data scarcity situation.
As the amount of data increases as in cases (J) and (M), the quality of the generated image also increases, producing good-quality images across all domains.


\begin{table}[]
\begin{center}
\begin{minipage}[t]{1.0\linewidth}
\centering
\caption{Results of the image to image translation method, StarGAN v2, in FID.}
\label{tab:fid_i2i1}
\footnotesize{
\begin{tabular}{ |c|c|c|c|c|c|c|c|c|c|c|} 
\hline
\multicolumn{1}{|c|}{} & \multicolumn{7}{|c|}{StarGAN v2}       \\ 
\hline
\multicolumn{1}{|c|}{} & \multicolumn{2}{|c|}{AfHQ} &\multicolumn{3}{|c|}{CelebAMask-HQ} &\multicolumn{2}{|c|}{Animal Faces} \\ 
\hline
  &(C) &(D) &(H) &(I) &(J) & (K) & (L)   \\ 
\hline
Lat. guided & 383.49 &338.75 &279.91 &182.65 &35.07 & 365.05& 172.44  \\
Ref. guided & 161.40 &237.74 &192.02 &86.94 &36.78 & 345.09& 155.02  \\ 
\hline

\end{tabular}
}
\end{minipage}
\begin{minipage}[t]{1.0\linewidth}
\centering
\caption{Results of the image to image translation method, FUNIT, in FID.}
\label{tab:fid_i2i_FUNIT}
\footnotesize{
\begin{tabular}{ |c|c|c|c|c|c|c|c|c|c|c|} 
\hline
\multicolumn{1}{|c|}{} &\multicolumn{7}{|c|}{FUNIT}  \\
\hline
\multicolumn{1}{|c|}{}  &\multicolumn{7}{|c|}{Animal Face} \\
\hline
& \multicolumn{2}{|c|}{(K)} &\multicolumn{3}{|c|}{(L)} &\multicolumn{2}{|c|}{(M)} \\
\hline
Lat. guided  & \multicolumn{2}{|c|}{-} &\multicolumn{3}{|c|}{ -} &\multicolumn{2}{|c|}{ -} \\
Ref. guided  &\multicolumn{2}{|c|}{280.09} &\multicolumn{3}{|c|}{163.06} &\multicolumn{2}{|c|}{33.48} \\
\hline
\end{tabular}
}
\end{minipage}
\end{center}
\end{table}
\figepstarganceleba
\figepstarganafhq
\figepstargananimals
\figepfunit

We evaluated FUNIT \cite{Liu2019FewShotUI} under a different scheme than other benchmark networks. FUNIT was originally proposed to target the few-shot scenario, where there is only a handful of data (e.g., 5 or 10 images) available in each class. The authors designed the network to learn generalizable appearance patterns from abundant amount of classes during the training phase while images in each class is limited. 
We further stressed the proposed model with setting (K) and (L), where the number of available classes and images per class are largely reduced. \autoref{tab:fid_i2i_FUNIT} shows the quantitative results in FID and the qualitative results are presented in \autoref{fig:epfunit}. It can be observed that reducing the number of classes (setting (L)) has visible impact on the performance, the trained model failed on capturing and transferring the target appearance. When the number of image per class is reduced along with the number of classes as in setting (K), the trained model even delivered visually unrealistic results.
Despite the promising performance provided by FUNIT when trained on the full dataset (setting (M)), we believe that there remains room for improvement in this line of research because abundant amount of classes are not always available in real-world  scenarios.



\section{Conclusion}\label{sec7}
In this work, we explored generative adversarial networks (GANs) for image synthesis and analyzed the performance of state-of-the-art architectures when working with limited datasets.
Firstly, the work included a concise overview of the GAN fundamentals and focused on analyzing state-of-the-art methods for different types of applications. We focused then on effective strategies for dealing with limited data, including data augmentation techniques and latent space analysis using GAN inversion techniques. Finally, the most commonly used metrics for performance evaluation were analyzed.

In the second part, we trained some widely-used architectures in different data scarcity regimes and evaluated their performance. 
The experimental analysis showed the level of voracity of the architectures and how many of them suffer from mode collapse problems in the presence of limited data, generally failing to achieve a sufficient level of image quality. Among the observed architectures, those of the semantic image synthesis task were the ones able to achieve the best results from a quantitative and qualitative point of view, even using only a few dozen training images.
We recognize that new stable diffusion models are capable of achieving better performance in absolute terms on the quality of the synthesized image. However, they are also extremely dependent on large amounts of data and even higher computational resources and training times. 

In real-world scenarios such as visual quality inspection or the medical field, the availability of sufficient data often becomes a significant obstacle. The main objective of this study is to highlight the challenges faced in data-driven generative approaches and to support the development of new methods that rely less on data.

\backmatter

%
\bmhead{Acknowledgments}
The work was supported by Baden-W\"urttemberg Ministry of Economic Affairs, Labour, and Tourism under the grant 017-180036 (project KI-Fortschrittszentrum ``Lernende Systeme und Kognitive Robotik").


\bigskip




\clearpage


\end{document}